\documentclass{article}
\usepackage{PRIMEarxiv}

\usepackage[utf8]{inputenc} 
\usepackage[T1]{fontenc}    
\usepackage{hyperref}       
\usepackage{url}            
\usepackage{booktabs}       
\usepackage{amsfonts}       
\usepackage{nicefrac}       
\usepackage{microtype}      
\usepackage{fancyhdr}       
\usepackage{graphicx}       

\usepackage{color}
\usepackage{mdframed} 
\usepackage{xcolor}
\usepackage{amssymb}   
\usepackage{titlesec}
\usepackage{glossaries}

\usepackage{tabularx,multirow}
\usepackage{times}
\usepackage{epsfig}
\usepackage{nicefrac}

\usepackage{cite}
\usepackage{amsmath,amssymb}
\usepackage{algorithmic}
\usepackage{algorithm}

\title{Blending 3D Geometry and Machine Learning for Multi-View Stereopsis}

\author{Vibhas K. Vats$^1$,
Md. Alimoor Reza$^2$,
David Crandall$^1$,
Soon-heung Jung$^3$ \\
$^1$ Luddy School of Informatics, Computing, and Engineering, Indiana University, Bloomington, IN 47408, USA\\
$^2$ Department of Mathematics and Computer Science, Drake University, Des Moines, IA 50311, USA\\
$^3$ Electronics and Telecommunications Research Institute, Daejeon 34129, Republic of Korea}

\begin{document}
\maketitle

\begin{abstract}
Traditional multi-view stereo (MVS) methods primarily depend on 
photometric and geometric consistency constraints. 
In contrast, modern learning-based algorithms often rely 
on the plane sweep algorithm to infer 3D geometry, 
applying explicit geometric consistency (GC) checks 
only as a post-processing step, with no impact on the 
learning process itself. In this work, we introduce \emph{GC-MVSNet++}, 
a novel approach that actively enforces geometric consistency of 
reference view depth maps across multiple source views (multi-view) 
and at various scales (multi-scale) during the learning phase (see Fig. \ref{fig:gc-module-idea}). 
This integrated GC check significantly accelerates the learning 
process by directly penalizing geometrically inconsistent pixels, 
effectively halving the number of training iterations 
compared to other MVS methods. Furthermore, we introduce a densely connected 
cost regularization network with two distinct block 
designs—simple and feature-dense—optimized to harness dense 
feature connections for enhanced regularization. 
Extensive experiments demonstrate 
that our approach achieves a new state-of-the-art on the BlendedMVS dataset, and 
competitive performance on the 
DTU and Tanks and Temples benchmark. To our knowledge, 
\emph{GC-MVSNet++} is among the few approaches that enforce supervised 
geometric consistency across multiple views and at multiple scales during training. 
Our code is available 
at \emph{https://github.com/vkvats/GC-MVSNet-PlusPlus}
\end{abstract}



\section{Introduction}
\label{sec:intro}

Traditional multi-view stereo (MVS) methods, such as 
Gipuma~\cite{Galliani2015fusibile}, Furu~\cite{Furukawa2010AccurateDA}, COLMAP~\cite{Johannes2016pixelwise}, 
and Tola~\cite{tola2011largescale},  primarily focused on addressing photometric 
and geometric consistency constraints across multiple views. 
In contrast, recent machine learning-based MVS approaches
\cite{ding2022transmvsnet,peng2022rethinkingMVS, Luo2019Pmvsnet, yao2019recurrent,chen2019pointbased, gu2019casmvsnet, yang2019CVPMVS,Cheng2019USCNet, wei2021aa, xu2019multiscale, yao2018mvsnet,YU2021AAcostvolume, wang2024mvssurvey, ma2025confident}
 have used deep networks 
to extract feature maps and combine them into 3D cost volumes, 
capturing subtle similarities between features. 
These advances—ranging from multi-level feature 
extraction~\cite{ding2022transmvsnet, gu2019casmvsnet} 
and attention-driven feature matching~\cite{ding2022transmvsnet, vats2023gcmvsnet} to 
refined cost-volume creation~\cite{ding2022transmvsnet,peng2022rethinkingMVS,yao2019recurrent,vats2023gcmvsnet} 
and innovative loss 
formulations~\cite{ding2022transmvsnet,peng2022rethinkingMVS,wei2021aa}—have significantly improved the fidelity of depth 
estimates and point cloud reconstructions.

However, unlike the traditional approaches, 
these machine learning-based techniques typically
model geometric consistency in only an indirect way. 
Typically, consistency checks 
are reserved for post-processing~\cite{ding2022transmvsnet, yao2019recurrent, gu2019casmvsnet, YU2021AAcostvolume}, 
filtering depth maps only 
after inference is complete. This means the rich, multi-view 
geometric constraints are not explicitly modeled during  
learning, leaving the network to uncover these 
relationships on its own, often in subtle and less direct ways.

\begin{figure*}[t]
\begin{center}
    \includegraphics[width=1\textwidth,height=12\baselineskip]{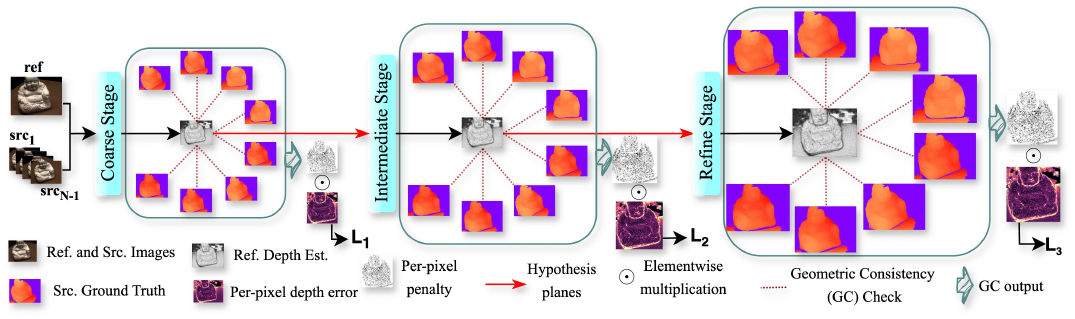}
    \caption{Our multi-view, multi-scale geometric consistency 
    process. It explicitly models the geometric consistency of the 
  estimated depth map across multiple source views during training at
  coarse, intermediate, and refine stages. At each stage, the ref. depth estimate map undergo forward-backward-reprojection 
  (see Alg. \ref{alg:forward-backward-reprojection}) across $N$ src. views 
  to generate the GC-penalty map to penalize per-pixel depth error. 
  This approach enables the model to learn geometric constraints 
  more quickly and accurately, leading to improved reconstruction 
  quality during inference.}
  \label{fig:gc-module-idea}
\end{center}
\end{figure*}

In this paper, we demonstrate that incorporating explicit multi-view geometric cues 
through geometric consistency checks across multiple source views 
during training (see Fig. \ref{fig:gc-module-idea}) significantly 
enhances accuracy while reducing the number of training iterations. 
To further improve the MVS pipeline, we present a novel densely 
connected cost regularization network with two distinct block architectures, 
specifically designed to fully exploit dense feature connections 
during the cost regularization process. 
Additionally, we enhance the feature extraction network 
with weight-standardized deformable convolution, 
ensuring improved feature extraction.

These innovations lead to the development of \emph{GC-MVSNet++}, 
a multi-stage model that learns geometric cues across three scales. 
At each scale, we integrate a  multi-view geometric 
consistency module that performs geometric consistency checks 
on reference view depth estimates across multiple source views, 
generating a per-pixel penalty. This penalty is then combined 
with the per-pixel depth error, calculated using cross-entropy 
loss at each stage, to form the final loss function, guiding 
the model toward more precise reconstructions.

This formulation of the loss function offers richer geometric cues that significantly expedite model learning. 
Our extensive experiments reveal that \emph{GC-MVSNet++} nearly halves 
the number of training iterations compared to other contemporary 
models \cite{ding2022transmvsnet, peng2022rethinkingMVS, gu2019casmvsnet, wei2021aa, yao2018mvsnet}. 
Our approach not only sets a new benchmark for accuracy on the BlendedMVS \cite{yao2019blended} dataset, 
but also secures 
the second position on the DTU \cite{jensen2014dtu} and Tanks and Temples \cite{Knapitsch2017tnt} advanced benchmark. 
GC-MVSNet++ is novel in its use of multi-view, multi-scale geometric consistency 
checks during training with a novel cost regularization network. Extensive ablation experiments 
further underscore the efficacy of our proposed method. To summarize:

\begin{itemize}
    \item[$\bullet$] We introduce an innovative multi-view, multi-scale geometric consistency module that fosters geometric coherence throughout the learning process.
    \item[$\bullet$] Our method reduces the training iterations by almost 50\% compared to other models, thanks to its explicit geometric cues.
    \item[$\bullet$] The versatility of GC-module allows it to be seamlessly integrated into various MVS pipelines to boost geometric consistency during training.
    \item[$\bullet$] Additionally, the GC module serves as a pre-processing tool to eliminate geometrically inconsistent pixels from the ground truth data.
    \item[$\bullet$] We also present a novel cost regularization network featuring two distinct block designs that leverage dense connections for cost regularization.
\end{itemize}

\noindent A preliminary conference version of this work appeared in  \cite{vats2023gcmvsnet}.
In this updated version, we introduce an 
innovative cost regularization network (Sec. \ref{subsec:costregnet}) and employ
depth filtering (Sec. \ref{subsec:depth-filtering}) to enhance GC-MVSNet and achieve better experimental results.
We also offer additional architectural details and
perform more extensive experimental 
evaluations.

\section{Related Work}\label{sec:related works}

Furukawa and Ponce \cite{Furukawa2010AccurateDA} propose a taxonomy 
that classifies MVS methods into four primary scene representations: 
\textit{volumetric fields} \cite{kutulakos1999spacecarving, seitz1997photorealistic, faugeras1998, sinha2007graphcut}, 
\textit{point clouds} \cite{chen2019pointbased, lhuillier2005quasidense}, 
\textit{3D meshes} \cite{fua1995object}, 
and \textit{depth maps} \cite{Galliani2015fusibile, Johannes2016pixelwise, ding2022transmvsnet, peng2022rethinkingMVS, gu2019casmvsnet, Cheng2019USCNet, xu2019multiscale, yao2018mvsnet, YU2021AAcostvolume, Campbell2008UsingMH, ma2025confident, guo2025murre}. 
Depth map-based methods can be further divided into traditional 
techniques that rely on feature detection and geometric constraint 
solving \cite{Galliani2015fusibile, Furukawa2010AccurateDA, Johannes2016pixelwise, Campbell2008UsingMH}, 
and learning-based approaches \cite{ding2022transmvsnet, peng2022rethinkingMVS, gu2019casmvsnet, Cheng2019USCNet, xu2019multiscale, yao2018mvsnet, YU2021AAcostvolume}, 
which have gained significant popularity in recent years.

Among the learning-based techniques, MVSNet \cite{yao2018mvsnet} presents 
a single-stage MVS pipeline that encodes camera parameters through 
differential homography to construct 3D cost volumes. 
However, it demands substantial memory and computational 
resources due to its use of 3D U-Nets \cite{olaf2015unet} 
for cost volume regularization. To address this challenge, 
subsequent research has pursued two primary strategies: 
the adoption of recurrent neural networks (RNNs) \cite{yao2019recurrent, wei2021aa, yan2020dynamicfusion, xu2021nonlocalrecurrent} 
and the development of coarse-to-fine multi-stage methods \cite{ding2022transmvsnet, peng2022rethinkingMVS, gu2019casmvsnet, Cheng2019USCNet, xu2019multiscale}.

\begin{figure*}[t]
\begin{center}
    \includegraphics[width=1\textwidth, height=14\baselineskip]{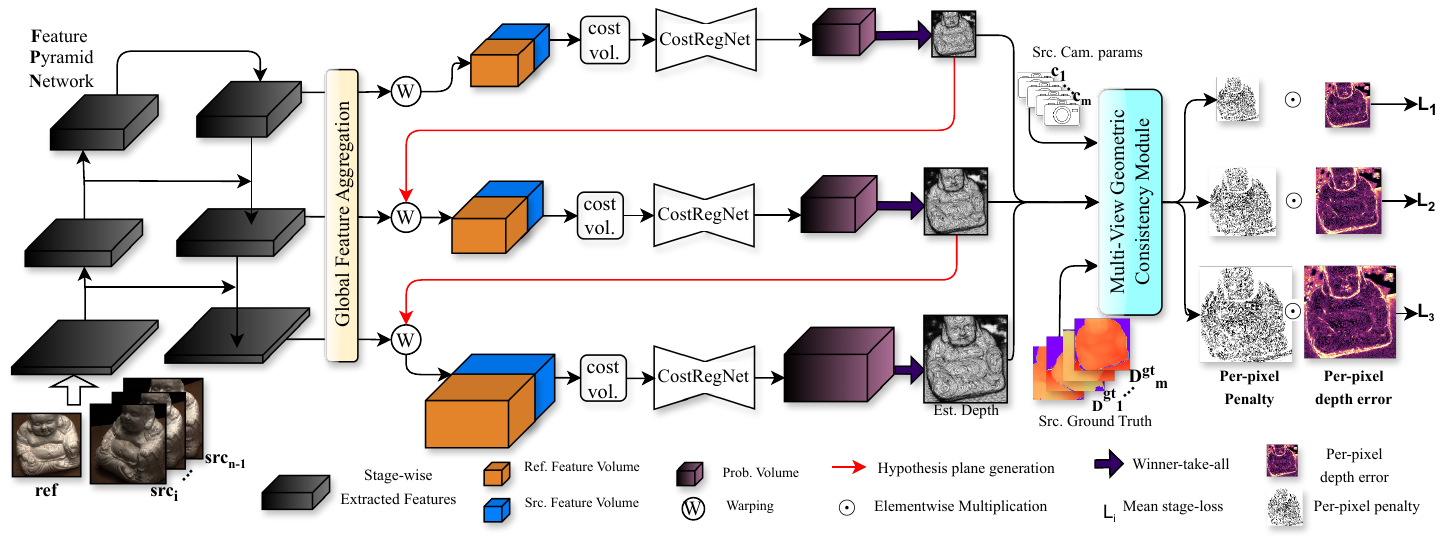}
    \caption{The GC-MVSNet++ architecture. It incorporates the GC module at the end of each stage. 
    This module utilizes the estimated reference view depth, $M$ source view ground truths, 
    and their associated camera parameters to conduct a multi-view geometric consistency check.
 It generates a per-pixel penalty ($\xi_p$) for reference view, 
 which is then element-wise multiplied with per-pixel depth error ($\xi_d$) to compute the stage loss $L_i$. $\xi_d$ is calculated using cross-entropy loss. 
 Figs. \ref{fig:dense-cost-reg-net} and \ref{fig:feature-pyramid-network} provide a detailed view of the Cost regularization network and feature extraction network, respectively.}
    \label{fig:gc-mvsnet architecture}
\end{center}
\end{figure*}

Coarse-to-fine multi-stage methods 
\cite{ding2022transmvsnet, peng2022rethinkingMVS, Luo2019Pmvsnet, yao2019recurrent, chen2019pointbased, gu2019casmvsnet, yang2019CVPMVS, Cheng2019USCNet, wei2021aa} 
have markedly enhanced the quality of depth estimates and point 
cloud reconstructions. These methods start by predicting a 
low-resolution (coarse) depth map and then refine it progressively. 
For instance, CasMVSNet \cite{gu2019casmvsnet} extends the 
single-stage MVSNet \cite{yao2018mvsnet} into a multi-stage framework, 
while TransMVSNet \cite{ding2022transmvsnet} improves performance 
over CasMVSNet through advanced feature matching techniques. 
UniMVSNet \cite{peng2022rethinkingMVS} further builds 
on CasMVSNet by employing a unified loss formulation 
to enhance accuracy. CVP-MVSNet \cite{yang2019CVPMVS} 
constructs a cost volume pyramid in a coarse-to-fine manner, 
and UCS-Net \cite{gu2019casmvsnet} introduces an adaptive 
thin volume module that optimally partitions the local 
depth range with fewer hypothesis planes. Additionally, 
TransMVSNet \cite{ding2022transmvsnet} incorporates 
transformer-based feature matching \cite{vaswani2017attention, angelos2020linearattention} 
to improve feature similarity, while UniMVSNet \cite{peng2022rethinkingMVS} 
integrates the benefits of both regression and classification 
approaches through a unified focal loss within its multi-stage framework.

While these methods enhance multi-stage MVS pipelines by refining specific components, 
they generally lack explicit integration of multi-view geometric 
cues during the learning process. Consequently, 
these models depend on the limited geometric information 
from multiple source views and the cost function 
formulation during training. 
Xu and Tao \cite{xu2019multiscale} 
address this limitation with a multi-scale geometric 
consistency-guided MVS approach that uses multi-hypothesis 
joint view selection to leverage structured region 
information for improved candidate hypothesis sampling. 
They argue that upsampled depth maps from source images 
can impose geometric constraints on estimates and utilize 
reprojection error \cite{Johannes2016pixelwise, zhang2008consistentvideo, vats2025Geometry} 
to assess consistency. Unsupervised methods~\cite{dai2019mvs2, truong2021warp} use
cross-view consistency to learn geometric features.
In contrast, our method employs forward-backward reprojection across 
multiple source views to directly evaluate and enforce 
geometric consistency in depth estimates, 
generating per-pixel penalties for geometrically inconsistent pixels. 

\noindent \textbf{Cost volume regularization:}
Cost volume regularization networks enhance raw cost volumes by converting 
them into regularized (smooth) volumes prior to depth estimation. 
Early approaches utilized the 3D-UNet architecture for similar 
regularization in stereo methods \cite{kendallcvpr17, changcvpr18, Zhang2019GANet}. 
These techniques typically employed variations of a 
basic encoder-decoder network. For instance, 
PSMNet \cite{changcvpr18} introduced a three-tier Hourglass architecture, 
while GANet \cite{Zhang2019GANet} incorporated a cost aggregation block 
featuring semi-global and local guided modules. 
Additionally, Guo et al. \cite{guo2019group} combined correlation 
and cost volume methods to leverage their respective strengths.

Building on its success in stereo and MVS tasks \cite{kendallcvpr17, Ji2017surfacenet, kar2017learning}, 
MVSNet applies a multi-scale 3D-UNet for cost volume regularization. 
R-MVSNet \cite{yao2019recurrent} enhances this with a recurrent GRU model to refine the cost volumes. 
CasMVSNet \cite{gu2019casmvsnet} uses separate networks for each stage, 
guiding each subsequent stage with the depth estimate from the previous one. 
PointMVSNet \cite{chen2019pointbased} further extends this approach to four stages. 
However, many recent methods \cite{ding2022transmvsnet, peng2022rethinkingMVS, gu2019casmvsnet, wei2021aa, vats2023gcmvsnet, zhe2023geomvsnet} 
do not significantly innovate in the architecture of cost regularization networks, 
typically employing simple encoder-decoder structures with 3D convolutional layers. 
In contrast, we propose an advanced densely connected U-Net architecture featuring 
two distinct block designs for the encoding and decoding phases.

\begin{algorithm}[t]
\footnotesize
\begin{algorithmic}
    \STATE \textbf{Inputs:} $D_0, c_0, D^{gt}_{i}, c^{gt}_{i}, D_{pixel}, D_{depth}$
    \STATE \textbf{Output:} $per\_pixel\_penalty$
    \STATE \textbf{Require} $M \ge N$
    \STATE $mask\_sum \gets 0$
    \STATE $D \gets {D_1^{gt}, ...D_M^{gt}}$
    \STATE $c \gets {c_1^{gt}, ...c_M^{gt}}$
    \FOR{$D_i^{gt},c_i^{gt}$ in $zip(D,c)$} 
    \STATE $D''_{P''_{0}}, P''_{0} \gets FBR(D_0,c_0,D_i^{gt},c_i^{gt})$ \hfill \COMMENT{Alg. \ref{alg:forward-backward-reprojection}}
    \STATE $PDE \gets {\vert\vert P_0 - P''_{0} \vert\vert_2 }$
    \STATE $RDD \gets {\frac{1}{D_0}\vert\vert D''_{P''_{0}} - D_0 \vert\vert_1}$
    \STATE $PDE_{mask} \gets {PDE > D_{pixel}}$
    \STATE $RDD_{mask} \gets {RDD > D_{depth}}$
    \STATE $mask \gets PDE_{mask} \lor RDD_{mask}$
    \IF{$mask > 0$}
        \STATE $mask \gets 1$ 
    \ELSE
        \STATE $mask \gets 0$
    \ENDIF
    \STATE $mask\_sum \gets mask\_sum + mask $
    \ENDFOR \\
$per\_pixel\_penalty \gets 1 + mask\_sum/M$
\end{algorithmic}
\caption{Geometric Consistency Check Algorithm}
\label{alg:gc-algorithm}
\end{algorithm}

\begin{algorithm}[t]
\footnotesize
\begin{algorithmic}
    \STATE \textbf{Inputs:} $D_0,c_0,D_i^{gt},c_i^{gt}$
    \STATE \textbf{Output:} $D''_{P''_{0}}, P''_{0}$
    \STATE $K_R, E_R \gets c_0$;  $K_S, E_S \gets c_i^{gt}$
    \STATE $D_{(R \rightarrow S)} \gets K_S \cdot E_S \cdot E_R^{-1} \cdot K_R^{-1} \cdot D_0$ \hfill\COMMENT{Project}
    \STATE $X_{D_{(R \rightarrow S)}}, Y_{D_{(R \rightarrow S)}} \gets D_{(R \rightarrow S)}$ 
    \STATE $D_{S_{remap}}\gets REMAP(D_i^{gt}, X_{D_{(R \rightarrow S)}}, Y_{D_{(R \rightarrow S)}})$ \hfill \COMMENT{Remap}
    \STATE $D''_{P''_{0}} \gets K_R \cdot E_R \cdot E_S^{-1} \cdot K_S^{-1} \cdot D_{S_{remap}}$   \hfill \COMMENT{Back project}
    \STATE $P''_{0} \gets (X_{D''_{P''_{0}}}, Y_{D''_{P''_{0}}})$
\end{algorithmic}
\caption{Forward Backward Reprojection (FBR)}
\label{alg:forward-backward-reprojection}
\end{algorithm}

\section{Methodology}\label{sec:methodology}

MVS methods take $N$ views as input, including a reference image $I_0 \in \mathbb{R}^{H\times W \times 3}$ and its paired ($N-1$) source view images $I^{N-1}_{i=1}$, along with the corresponding camera parameters $c_0, ..., c_N$,  and then   to estimate the reference view depth map ($D_0$) as the output. 

\subsection{Network Overview}\label{sec:network-overview}

The architecture of our approach Geometric Consistency MVSNet++, abbreviated as GC-MVSNet++, is shown in Fig. \ref{fig:gc-mvsnet architecture}.
We use a deformable convolution-based \cite{Dai2017deformableCNN}  feature pyramid network (FPN) \cite{Lin2016featurepyramid}
architecture (Sec. \ref{sec:other-modification}) to extract features
from input images in a coarse-to-fine manner in three stages. 
At only the coarse stage, we apply a 
Global Feature Aggregation (GFA) module with linear attention
\cite{angelos2020linearattention, ding2022transmvsnet} to leverage
global context information within and between reference and source view features.
At each stage, we build a correlation-based cost volume
of shape $H' \times W'\times D'_h \times 1$ using feature
maps of shape $N \times H' \times W' \times C$,
where $H'$, $W'$, and
$C$ denote the height, width, and number of channels of a given stage, and
$D'_h$ is the number of depth
hypotheses at the corresponding stage.
The cost volume is
regularized with the proposed dense-connected CostRegNet.
We use a
winner-takes-all strategy to estimate the depth
map $D_{0}$ at each stage. 

We employ the GC module at each stage. It
checks the geometric consistency of each pixel in $D_{0}$ across $M$
source views and generates $\xi_p$
(Sec. \ref{sec:multi-source-GC-module}), a pixel-wise 
factor that is
multiplied with the per-pixel depth error ($\xi_d$), calculated using a
cross-entropy function. It penalizes each pixel in $D_0$ for its
inconsistency across $M$ source views to accelerate geometric cues
learning during training. 
TransMVSNet \cite{ding2022transmvsnet} (without feature matching transformer and 
adaptive receptive field module) trained with cross-entropy loss is our baseline (GC-MVS-base), 
see Table \ref{table:stages-of-gcmvsnet++}.

\begin{figure}[t]
\begin{center}
   \includegraphics[width=0.6\linewidth]{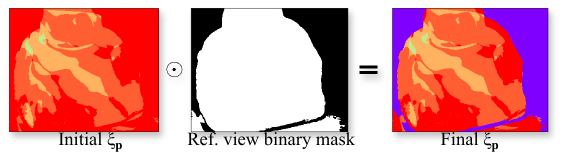}
    \caption{The final $\xi_p$ is the outcome of elementwise multiplication ($\odot$) of initial $\xi_p$ and reference view mask. It restricts the penalties within the reference view mask.}
    \label{fig:geo-weight-masking}
\end{center}
\end{figure}

\subsection{Multi-View Geometric Consistency Module}\label{sec:multi-source-GC-module}

\begin{figure*}[t]
\begin{center}
    \includegraphics[width=0.95\textwidth]{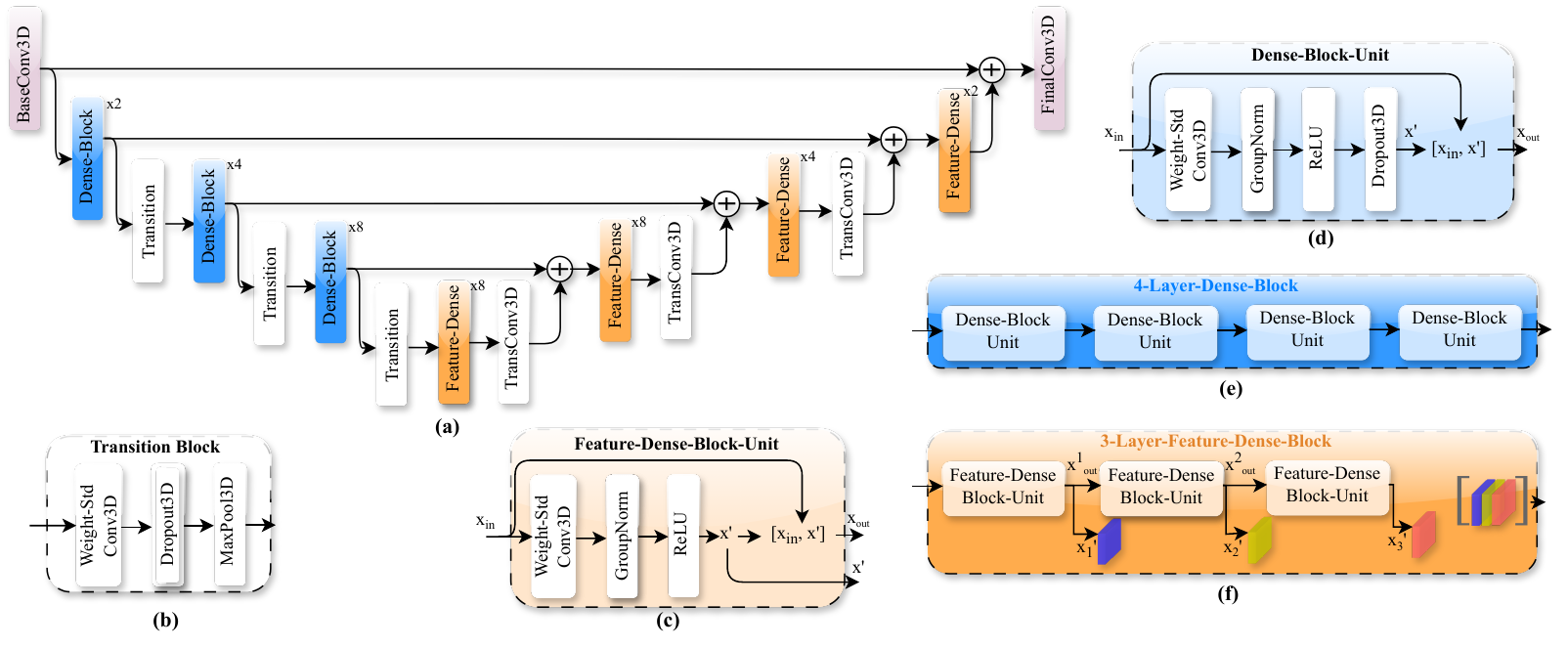}
    \caption{Expanded view of the proposed Cost Regularization Network (CostRegNet) from Fig. \ref{fig:gc-mvsnet architecture}. 
    (a) shows the architecture of the dense CostRegNet with three types of blocks -- 
    dense, feature-dense, and transition blocks. 
    (b) shows the internal architecture of the transition block, 
    its components include a weight-standardized
    3D convolution layer followed by a 3D dropout and 3D max-pooling layers. 
    (c) and (d) show a single unit inside the feature dense and dense units. 
    (e) shows a four-layer dense block and (f) shows a three-layer feature dense block.}
    \label{fig:dense-cost-reg-net}
\end{center} 
\end{figure*}

GC-MVSNet++ estimates reference depth maps at three stages with
different resolutions. At each stage, the GC module takes $D_0$, $M$
source view ground truths $D^{gt}_1, ... D^{gt}_{M}$, and their camera
parameters $c_0, ... c_M$ as input (see Alg. \ref{alg:gc-algorithm}). The GC module is then initialized
with a \textit{geometric inconsistency mask sum} (or $mask\_sum$) of
zero at each stage. This mask sum accumulates the inconsistency of
each pixel across the $M$ source views. For each source view, the GC
module performs \textit{forward-backward reprojection}  of $D_0$
to generate the penalty and then adds it to the mask sum.

Forward-backward reprojection (FBR), as shown in Alg. \ref{alg:forward-backward-reprojection}, is a crucial three-step process. First,
we project each pixel $P_{0}$ of $D_{0}$ to
its $i^{th}$ neighboring source view using intrinsic ($K_R, K_S$) and extrinsic ($E_R, E_S$) camera 
parameters to obtain corresponding pixel $P'_{i}$, and denote the corresponding depth map as $D_{(R \rightarrow S)}$. Second, we 
similarly remap $D_i^{gt}$ 
to obtain $D_{S_{remap}}$.
Finally, 
we back project $D_{S_{remap}}$ to the reference view using intrinsic and extrinsic camera parameters 
to obtain $D''_{P''_{0}}$ (see Alg. \ref{alg:forward-backward-reprojection}). $D_0$ and $D''_{P''_{0}}$ represent
the depth values of pixels $P_0$ and
$P''_{0}$ \cite{Hartley2012Multi-view-geometry}. With $P''_{0}$
and $D''_{P''_{0}}$, we calculate the pixel displacement error (PDE)
and relative depth difference (RDD). PDE is the $L_2$ norm between
$P_0$ and $P''_{0}$ and RDD is the absolute value difference between
$D''_{P''_{0}}$ and $D_0$ relative to $D_0$ as shown in Alg. \ref{alg:gc-algorithm}.

For each stage, we generate two binary masks of inconsistent pixels, $PDE_{mask}$ and $RDD_{mask}$,
by applying thresholds
$D_{pixel}$ and $D_{depth}$, and then take a logical-OR of the two to produce a single mask
of inconsistent pixels.
These inconsistent pixels are assigned a value $1$ and all
other pixels, including the consistent and the out-of-scope pixels,
are assigned $0$ to form a penalty mask. This penalty mask
is then added to the mask sum (Alg. \ref{alg:gc-algorithm}),
which accumulates the penalty mask for each of the $M$ source views to generate a final
mask sum with values $\in [0, M]$.
Each pixel value indicates the number of inconsistencies of the pixel
across the $M$ source views.

From this mask sum, we then generate the inconsistency penalty $\xi_p$
for each pixel. Our initial approach generated $\xi_p$ by dividing the mask sum by 
$M$ to normalize within the $[0,1]$.
However, we found that using $\xi_p$ itself for element-wise
multiplication reduces the contribution of perfectly consistent (zero
inconsistency) pixels to zero, preventing further improvement of such
pixels.  To avoid this, we add $1$ so that elements of $\xi_p \in [1,2]$. 
A reference view binary mask is applied on initial $\xi_p$ to
generate the final $\xi_p$, as shown in Fig. \ref{fig:geo-weight-masking}.

\noindent \textbf{Occlusion and its impact:}
In multi-view stereo, occluded pixels naturally occur 
as 3D points often remain invisible from certain views. 
These hidden pixels significantly affect geometric constraints, 
leading to the penalization of reference view pixels corresponding 
to occluded 3D points as inconsistent. To prevent these occluded pixels 
from disproportionately impacting geometric consistency losses, 
careful management is essential.

Although some methods explicitly model occlusion 
\cite{kang2001occlusion, nakamura1996stereoocclusion}, 
our approach is inherently resilient to occlusion due to three key factors. 
First, we select the closest $M$ source views, as defined in 
MVSNet \cite{yao2018mvsnet}, to reduce the occurrence of occluded pixels 
across different views. During the FBR process, we remap $D_i^{gt}$ to 
generate $D_{S_{remap}}$ and perform back projection 
as detailed in Alg. \ref{alg:forward-backward-reprojection}. 
This remapping and back projection effectively manage severe 
occlusions (refer to Appendix A in Supplemental Material). 
Lastly, we use a binary mask on $\xi_p$, as illustrated 
in Fig. \ref{fig:geo-weight-masking}, 
to confine penalties to valid reference view pixels only. 
These combined strategies enable us to address the 
challenges posed by occluded pixels and prevent the explosion of loss values.

\begin{figure*}[t]
\begin{center}
    \includegraphics[width=0.9\textwidth,height=14\baselineskip]{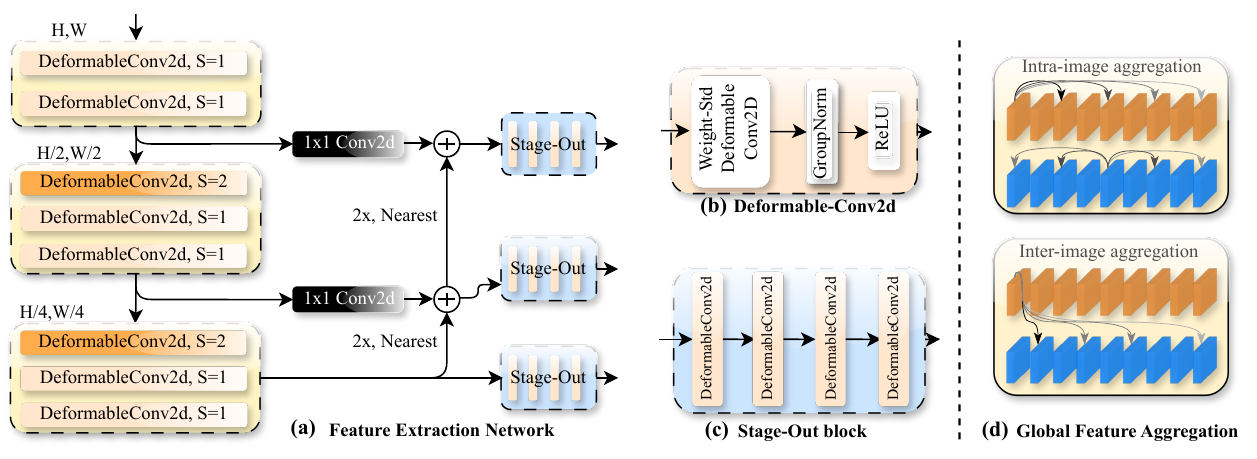}
    \caption{Expanded view of the Deformable Feature Extraction Network (FEN) and its 
    components. (a) shows the overall architecture of the FEN. (b) and (c) shows 
    the internal design of the deformable-conv2d and stage-out block. (d) shows 
    the two distinct methods of global feature aggregation for reference and 
    source image features.}
    \label{fig:feature-pyramid-network}
\end{center} 
\end{figure*}

\subsection{Cost Regularization Network (CostRegNet)} \label{subsec:costregnet}

The raw cost volume derived from reference and source image features 
often contains significant noise, originating from occlusions, 
feature mismatches, and non-Lambertian surfaces. 
This noise can hinder the accuracy of depth estimation, 
necessitating regularization to achieve smoother results. 
To address this, we introduce a novel densely connected~\cite{densenet_CVPR17} 
cost regularization network, termed dense-CostRegNet, 
designed specifically for dense prediction tasks.

Dense-CostRegNet leverages the DenseNet \cite{densenet_CVPR17} architecture, 
known for its dense block connections. 
DenseNet was developed as an enhancement over ResNet \cite{resnet_cvpr16}, 
focusing on improved image recognition through its dense connectivity within blocks. 
Drawing inspiration from its success in image recognition, 
we have adapted this concept into a U-Net \cite{olaf2015unet} 
style encoder-decoder framework. Our design features distinct 
encoder and decoder blocks tailored for depth 
estimation challenges, as illustrated in Fig. \ref{fig:dense-cost-reg-net} (a).

In the U-Net architecture, the encoding phase employs a 
\textit{simple} dense block, closely adhering to the original DenseNet design. 
In contrast, the bottleneck and decoder stages utilize a \textit{feature-dense} block. 
This feature-dense block emphasizes generating new features 
while capitalizing on the benefits of dense feature 
connections within the block, as depicted in Fig. \ref{fig:dense-cost-reg-net} (c) and (f). 
We believe that the dense connections within these blocks are crucial for enhancing the 
smoothness of the cost volume, which leads to more accurate depth estimates. 
Below, we provide a detailed explanation of each component of the dense-CostRegNet, 
beginning with an overview of dense connectivity and its advantages.

\subsubsection{Dense Connectivity}
Let us assume $H_{k}(.)$ denotes the feed-forward 2D convolution operation in a traditional 
2D convolutional neural network (CNN), which feeds the output feature map $x_{k-1}$ of 
the $(k-1)^{th}$ layer to the input of the next layer, $x_{k} = H_{k-1}(x_{k-1})$. 
To bypass the non-linear transformation, ResNet~\cite{resnet_cvpr16} additionally adds  
skip-connections, $x_{k} = H_{k-1}(x_{k-1}) + x_{k-1}$. DenseNet~\cite{densenet_CVPR17} 
further improves the information flow between layers by adding a denser connectivity 
pattern with direct connections from any layer to all its subsequent layers. 
More precisely, the $k^{th}$ layer receives the concatenation of 
feature maps [$x_{0}, x_{1}, ...x_{k-1}$] from all its preceding layers,
\begin{equation}
    x_{k} = H([x_{0}, x_{1}, ...x_{k-1}]),
    \label{eq:dense-connection}
\end{equation}
$H_{k}(.)$ is a composite of sequential layer operations. 
The output feature dimension of the $k^{th}$ layer depends on $g$ 
(a fixed \textit{growth rate} parameter). This can be easily extended 
to 3D convolutional layers.

\subsubsection{Encoder Dense Blocks}

The encoder uses a 3D version of the simple formulation of a dense block
as discussed in Eq. \ref{eq:dense-connection}. It reduces the feature 
resolution to $\frac{1}{8}$ of the input to reach the bottleneck. 
Each stage of the encoder (with fixed feature resolution)
has only one dense block but uses a different number 
of dense-block units. 
Fig. \ref{fig:dense-cost-reg-net} (c) shows the design 
of a single dense-block unit with weight-standardized 3D convolution, followed 
by a group normalization layer, ReLU non-linearity, and a 3D dropout layer. The 
learned features ($x'$) are concatenated with the input feature ($x_{in}$) for
the next dense-block unit. Fig. \ref{fig:dense-cost-reg-net} (e) shows a 4-layer
dense block. The number of layers (dense units) inside each block is shown as 
a superscript in the U-Net diagram (Fig. \ref{fig:dense-cost-reg-net}(a)).

\subsubsection{Decoder Dense Blocks}
The bottleneck and the decoder parts of the U-Net use the same feature-dense block. 
Fig. \ref{fig:dense-cost-reg-net} (c) shows the internal design of a unit. It includes a 
weight-standardized 3D convolution followed by a GroupNorm \cite{he2018groupnorm} and ReLU 
non-linearity. The learned $g$ (dense block growth rate) 
new features $x'$ are concatenated with the input features 
($x_{in}$) and sent to the next feature-dense unit. These learned 
features from each unit ($x'_{1}, x'_{2}, ..., x'_{i}$) are kept aside
and concatenated at the end as the final output of the feature-dense blocks, 
see Fig. \ref{fig:dense-cost-reg-net} (f). This design has the same number of 
dense connections as the \textit{simple} dense blocks but only uses new features for the 
final output of the block. This allows subsequent layers to focus on new features while 
utilizing dense connectivity and encouraging better regularization of the 
cost volume. This design also makes it possible to
remove the dropout layer from it. Each feature-dense block follows 
a similar formulation as in Equation (\ref{eq:dense-connection}), but generates a 
different output,
\begin{equation}
    x_{out} = concat\left[x'_{0}, x'_{1}, ..., x'_{i-1}\right],
    \label{eq:efature-dense-connection}
\end{equation}
where subscript $i$ is the number of feature-dense units in the block. 
The number of feature-dense units for each block is shown as a superscript in the 
U-Net architecture (Fig. \ref{fig:dense-cost-reg-net} (a)). The encoder and the decoder 
dense blocks have the same number of basic units at the same level. The growth rate $g$ 
is fixed at $4$ throughout the network. 

\begin{figure*}[t]
\begin{center}
    \includegraphics[width=0.9\textwidth,height=10\baselineskip]{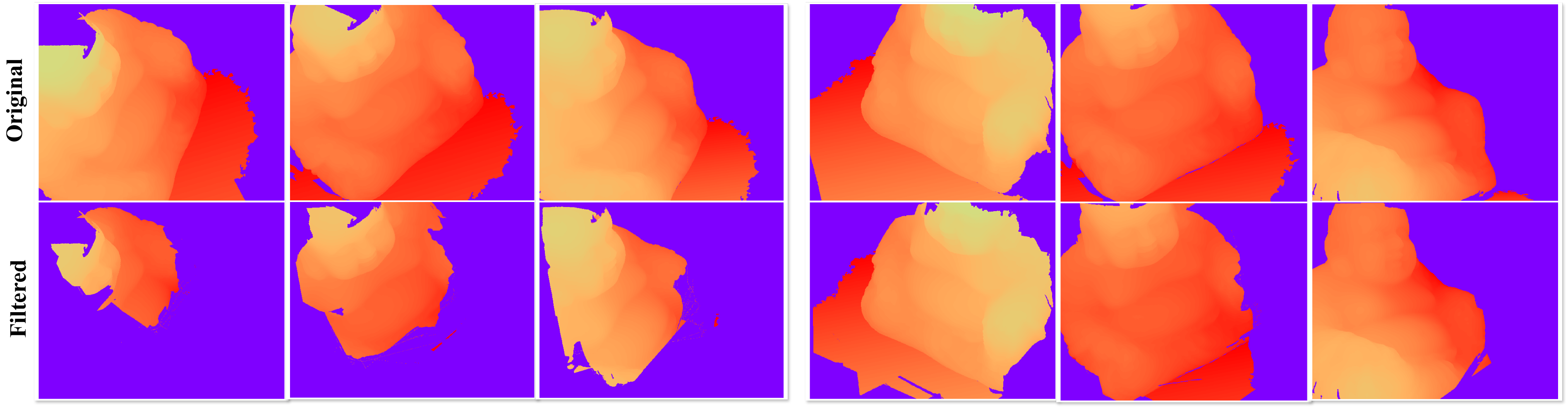}
    \caption{The visual comparison presents ground truth depth maps for scene 114 from the DTU dataset, showcasing various viewpoints before and after depth filtering. The top row displays the original ground truth depth maps from different viewpoints, while the bottom row illustrates the same maps following filtration. The left half of the figure highlights depth maps with the highest impact (error) from the filtration process, whereas the right half features depth maps with the lowest impact (error).}
    \label{fig:depth-filtering}
\end{center}
\end{figure*}

\subsubsection{Transition-down and Transition-up Blocks}
The feature maps are downsampled with \textit{transition blocks} to 
reduce the 3D spatial resolution (i.e., disparity, height, and width) of the 
feature maps. Fig.~\ref{fig:dense-cost-reg-net} (b) shows the architecture of 
the transition block. It is a three-layer block, a 1x1-3D convolution
followed by a dropout and a 3D max-pooling layer. Similar to the 
layers inside the dense blocks, the transition block also has 
a weight-standardized 3D convolution layer. Transition-up layers 
use a single weight-standardized transpose convolution layer to double 
the spatial resolution using 3x3x3 kernels. After each transition-up layer, 
the features from the same level are added together as shown in Fig. \ref{fig:dense-cost-reg-net}(a).

\subsection{Deformable Feature Extraction Network}\label{sec:other-modification}


Besides the geometric consistency module and dense-CostRegNet, we adopted a new architecture for 
the Feature Extraction Network (FEN) with weight standardized \cite{qiao2019weightstandardization}
deformation convolution layers \cite{Dai2017deformableCNN, zhu2018deformableV2}. 
Deformable layers are known to adjust their sampling locations based on model
requirements \cite{Dai2017deformableCNN, zhu2018deformableV2}. This
helps extract better features for accelerated learning.
We use a multi-level feature 
pyramid network design for semantically strong feature extraction \cite{Lin2016featurepyramid} 
at all levels.

Deformable convolutional blocks (Deformable-Conv2d) are the
building blocks of the FEN as shown in Fig. \ref{fig:feature-pyramid-network} (a).
Each deformable-conv2d block consists of three layers, a weight-standardized conv-2d layer, 
followed by a group normalization layer, and a ReLU activation. A total of eight such layers 
are used to extract image features for the three stages, $\frac{HW}{4}, \frac{HW}{2}$ and $HW$, 
as shown in Fig. \ref{fig:feature-pyramid-network}(a). 
FEN uses a strided deformable-conv2d layer instead of a max-pooling layer to 
reduce the feature map size at the start of a new stage. This helps to avoid direct feature 
loss by parameterless max-pooling compression. 
At the end of each resolution, a \textit{stage-out} block is applied 
to obtain the final output features. It consists 
of four deformable-conv2d layers, as shown in 
Fig. \ref{fig:feature-pyramid-network}(c). 
The lowest resolution features, $\frac{HW}{4}$, are the most semantically strong.
We upsample the feature maps (using 
 nearest neighbors) to propagate the strong semantic information 
to the higher-resolution stage. The upsampled features 
are then added to the features from the downsampling 
path at the same level. The lateral $1\times1$ convolution layer 
matches the number of channels of these two sets of features.  
The summed output is used as input for the \textit{stage-out} 
block, as shown in Fig. \ref{fig:feature-pyramid-network}(a). 

Unlike most MVS methods \cite{ding2022transmvsnet,peng2022rethinkingMVS, gu2019casmvsnet, wei2021aa, Zhang2020visibility,weilharter2021highresMVS} that use batch normalization 
\cite{ioffe2015batchnorm} during training, we use group normalization throughout our network.
As observed in \cite{ioffe2015batchnorm}, batch normalization provides more consistent
and stable training with large batch sizes, but it is inconsistent and
has a degrading effect on training with smaller batches. MVS
methods are restricted to very small batch sizes, often between $1-4$, due to
large memory requirements by the 3D-regularization network. Thus, we replaced batch normalization with 
group normalization layers
\cite{he2018groupnorm} of group size $4$ across the network. Group normalization performs normalization across a 
number of channels that is independent of the
number of examples in a batch \cite{he2018groupnorm}. We also implement weight standardization
\cite{qiao2019weightstandardization} for all layers (including 3D-convolutional layers) in the
network. With these modifications, we achieve stable and reproducible
training.

\subsection{Global Feature Aggregation}

Multi-view stereo matching is a one-to-many matching problem, which 
requires simultaneous consideration of all source views for effective matching. 
Following TransMVSNet \cite{ding2022transmvsnet}, we use the global 
feature aggregation (GFA) technique just before cost volume creation. GFA 
aggregates global context information using inter- and intra-image 
feature interactions \cite{ding2022transmvsnet, caomvsformer}, as shown in Fig. \ref{fig:feature-pyramid-network}(d). It has been 
proven  improve prediction quality and reduce matching 
uncertainties, especially for regions with little texture or repetitive patterns. 

GFA uses linear attention \cite{katharopoulos2020transformers} with 
multiple heads \cite{vaswani2017attention} to estimate attention scores
using \textbf{Q}uery and \textbf{K}ey, 
\begin{equation}
    Attention(Q, K, V) \, = \Psi(Q) (\Psi(K^{T})V),
\end{equation}
where $\Psi(.) = elu(.) + 1$,  $elu(.)$ represents 
the activation function of exponential linear unit \cite{ ding2022transmvsnet, clevert2015fast}, 
and \textbf{V} is the value in attention calculation.

The GFA module aggregates global features in two distinct manners: 
intra-image, in which 
the \textbf{Q}s and \textbf{K}s are from 
the same image (view), and inter-image, when they are
from different views.
Following TransMVSnet, both the 
source and  reference view features are updated during intra-image GFA, but  
only the source features are updated for inter-image GFA. 

The top of the Fig. \ref{fig:feature-pyramid-network}(d) shows intra-image aggregation. 
The first row shows the aggregation process for the 
reference features and the second row shows 
the aggregation for the source features. The darkness 
of the arrows  indicate the magnitude of 
attention. The bottom of Fig. \ref{fig:feature-pyramid-network}(d) 
shows inter-image aggregation 
where the features from the reference and
source images interact  to estimate the attention score for 
global feature aggregation. The aggregated reference and source view 
features are then used to create the cost volume.

\subsection{Cost Function}\label{sec:cost-function}

Most learning-based MVS methods \cite{gu2019casmvsnet, yang2019CVPMVS,Zhang2020visibility} 
treat depth estimation as a regression problem and use an $L_1$ loss between prediction and ground truth. Following AA-RMVSNet \cite{wei2021aa} and UniMVSNet \cite{peng2022rethinkingMVS}, we treat depth 
estimation as a classification problem and adopt a cross-entropy loss formulation 
from AA-RMVSNet \cite{wei2021aa} (see \cite{peng2022rethinkingMVS} for 
relative advantages of regression and classification approaches). The pixelwise 
depth error $\xi_d$ is calculated at each stage,
\begin{equation}
    \xi_d = \mathcal{D}(D^{gt}_{0}, D_0),
\end{equation}
where $D^{gt}_{0}$ is the reference ground truth and $D_0$
is the reference depth estimate. $\mathcal{D}$ denotes the cross-entropy function 
modified to produce per-pixel depth error between $D^{gt}_{0}$ and $D_0$.
We further
enhance the one-hot supervision by penalizing each pixel for its
inconsistency across different source views. This is implemented using
element-wise multiplication ($\odot$) between $\xi_d$ and $\xi_p$ at
each stage. The mean stage loss, $L_i$, is calculated as,
\begin{equation}
\begin{split}
    {L_i}_{(stage)} & = mean (\xi_p \odot \xi_d)\\
    \mathcal{L}_{total} & = \alpha.L_1 + \beta.L_2 + \gamma.L_3
\end{split}
\end{equation}
where ${L_i}_{(stage)}$ is the mean stage loss,  $\mathcal{L}_{total}$ is the total loss, and 
$\alpha, \beta$ and $\gamma$ are the stage-wise weights.
This formulation of the cost function
with pixel-level inconsistency penalties explicitly forces the model 
to learn to produce multi-view, geometrically-consistent depth maps.

\subsection{Depth Filtering}\label{subsec:depth-filtering}

In analyzing the penalty mask $\xi_p$, we observed that certain regions 
of the scene consistently receive penalties close to the maximum throughout 
training (see the final $\xi_p$ in Fig. \ref{fig:depth-filtering}). 
This contrasts sharply with other regions, which exhibit a rapid decrease 
in penalties as training progresses (lighter regions in the same figure). 
This discrepancy is closely tied to the methods used by MVSNet \cite{yao2018mvsnet} 
and R-MVSNet \cite{yao2019recurrent} for generating ground truth depth maps.

The DTU dataset \cite{jensen2014dtu} provides ground truth point clouds with normal information, 
which is employed in Screened Poisson Surface Reconstruction (SPSR) \cite{kazhdan2013screened} 
to create mesh surfaces. These surfaces are then rendered from each 
viewpoint to produce depth maps. We suspect that this 
ground truth generation process introduces geometric inconsistencies into the depth estimates. 
Consequently, since we use the source view ground truth to 
calculate per-pixel penalties $\xi_p$, this initial error is implicitly
fed into the model. This creates a cycle of incorrect feedback 
for geometric consistency checks, resulting in persistently high penalties for certain regions.

To mitigate the impact of inconsistent depth values, we apply the same GC module 
(refer to Alg. \ref{alg:gc-algorithm}) to filter accurate depth maps, with a minor adjustment. 
Specifically, we substitute the estimated reference depth map ($D_0$) 
with its ground truth counterpart ($D_{0}^{gt}$) in Alg. \ref{alg:gc-algorithm}. 
The hyperparameters are set to $D_{pixel}$=2, $D_{depth}$=0.25, and $M$=8, 
ensuring that only the most erroneous depth values are filtered out. 
Fig. \ref{fig:depth-filtering} displays the original 
and filtered depth maps from various viewpoints of a sample scene in the DTU dataset. 
The left half of the figure illustrates viewpoints 
with the most filtered-out values, while the right half shows viewpoints with the fewest.
Although we initially identified this depth map rendering issue in the DTU dataset, 
the approach is broadly applicable to other datasets employing similar methods 
for generating ground truth depth maps. For the BlendedMVS dataset, 
we use $D_{pixel}$=0.5, $D_{depth}$=0.05, and $M$=10 in the depth filtering process.


\begin{table*}[t]
\caption{Quantitative results on DTU and BlendedMVS. Accuracy (Acc), completeness (comp) and overall are in $mm$. We follow Darmon et al. \cite{darmon2021wildMVS} for BlendedMVS evaluation. \textbf{Bold} and \underline{underline} represents first and second place, respectively. 
Error maps are shown in Appendix H.}
\vspace{-10pt}
\setlength{\tabcolsep}{2pt}
  \begin{center}
    {\footnotesize{
\begin{tabular}{clccc|ccc}
\toprule
\multicolumn{5}{c}{\textbf{DTU Dataset}} & \multicolumn{3}{c}{\textbf{BlendedMVS Dataset}} \\
\midrule
& Method  & Acc $\downarrow$ & Comp $\downarrow$ & Overall $\downarrow$ & EPE $\downarrow$ & $e_1 \downarrow$ & $e_3 \downarrow$ \\
\cmidrule{2-8}
\parbox[t]{2mm}{\multirow{4}{*}{\rotatebox[origin=c]{90}{Traditional}}} & Furu \cite{Furukawa2010AccurateDA} & 0.613 & 0.941 & 0.777 & --  & --  & -- \\
& Tola \cite{tola2011largescale} & 0.342 & 1.190 & 0.766 & --  & --  & -- \\
& Gipuma \cite{Galliani2015fusibile}  & \textbf{0.283} & 0.873 & 0.578 & --  & --  & -- \\
& COLMAP \cite{Johannes2016pixelwise}  &  0.400 & 0.664 & 0.532 & --  & --  & -- \\ 
\cmidrule{2-8}
\parbox[t]{2mm}{\multirow{12}{*}{\rotatebox[origin=c]{90}{Learning-based}}}& MVSNet \cite{yao2018mvsnet}  & 0.396 & 0.527 & 0.462 & 1.49  &  21.98 &  8.32\\
& CasMVSNet \cite{gu2019casmvsnet}   & 0.325 & 0.385 & 0.355 &  1.43 & 19.01  &  9.77\\
& CVP-MVSNet \cite{yang2019CVPMVS}  & \underline{0.296} & 0.406 & 0.351 &  1.90 & 19.73  & 10.24  \\
& Vis-MVSNet \cite{Zhang2020visibility} & 0.369 & 0.361 & 0.365 & 1.47  & 15.14  &  5.13 \\
& AA-RMVSNet \cite{wei2021aa}  & 0.376 & 0.339 & 0.357 & --  & --  & -- \\
& EPP-MVSNet \cite{ma2021eppMVS} & 0.413 & 0.296 & 0.355 & 1.17  & 12.66 &  6.20 \\
& UniMVSNet \cite{peng2022rethinkingMVS}  & 0.352 & 0.278 & 0.315 & --  & --  & -- \\
& TransMVSNet \cite{ding2022transmvsnet}  & 0.321 & 0.289 & 0.305 & 0.73 & 8.32 & 3.62\\
& GBi-Net \cite{mi2022gbi} & 0.315 & 0.262 & 0.289 & --  & --  & -- \\
& MVSTER \cite{wang2022mvster} & 0.350 & 0.276 & 0.313 & --  & --  & -- \\
& GeoMVSNet \cite{zhe2023geomvsnet} &  0.331 & 0.259  & 0.295 & --  & --  & -- \\
& MVSFormer \cite{caomvsformer} & 0.327  & 0.251  & 0.289  & --  & --  & -- \\
& MVSFormer++ \cite{cao2024mvsformer} & 0.309  & 0.252  & \textbf{0.2805}  &  -- &  -- & -- \\
& GoMVS \cite{wu2024gomvs} & 0.347 & \textbf{0.227} & 0.287 & -- & -- & -- \\
& GC-MVSNet \cite{vats2023gcmvsnet} & 0.330 & 0.260 & 0.295 & \underline{0.48}  & 7.48  &  2.78\\
\cmidrule{2-8}
& \textbf{GC-MVSNet++} (ours) & 0.319 & \underline{0.246} & \underline{0.2825} & \textbf{0.407}  & \textbf{5.79}  &  \textbf{2.41}\\
\bottomrule
\end{tabular}
}}
\vspace{-6pt}
\label{Table:DTU-Blended-model-comparison}
\vspace{-15pt}
\end{center}
\end{table*}

\subsection{GC-MVSNet++ Design Insights}

The typical supervised MVS pipeline  has three interconnected parts: 
feature extraction, cost volume regularization, and the supervision signal. 
The feature extraction networks generate 
initial features used to form the initial cost volume, 
which is then refined through cost volume regularization to estimate depth. 
The learning process is driven by the supervision signal provided by the loss function.

Our method begins by enhancing supervision with geometric consistency, 
explicitly penalizing depth estimates lacking consistency across multiple views. 
This geometry-aware supervision encourages improved feature extraction 
with better physical understanding of the 3D scenes
and more accurate initial cost volume creation. 
However, the improved feature extraction and initial cost volume  
are limited by the rigidity of sampling locations in convolutional 
layers and the lack of global aggregation. 
To address this,  we first add a global feature aggregation module for 
 intra-image and inter-image feature fusion, and then 
adopt deformable 
convolution layers to allow flexible sampling 
that better accommodates geometric and view variations. 
(See Table \ref{table:stages-of-gcmvsnet++} for ablation experiments.)

Inspired by the benefits of flexible feature sampling, 
we sought a parallel form of flexibility within the cost volume regularization network. 
Dense network connectivity, as exemplified by DenseNet~\cite{densenet_CVPR17}, 
naturally supports flexible, multi-scale feature reuse with strong gradient flow 
for optimization~\cite{resnet_cvpr16, densenet_CVPR17, he2016identity}, 
thereby enhancing learning representational power. 
Leveraging this insight, we adopted dense connectivity 
inspired modules for increased feature reuse, 
stronger gradient flow, and improved landscape 
for optimization~\cite{hao2018visualizing, he2016identity}. 
While the encoding module (simple dense block) closely follows  a traditional DenseNet
for its enhanced encoding capabilities~\cite{densenet_CVPR17}, 
we designed a feature-dense decoding module 
specifically tailored for cost volume refinement. 
As illustrated in Fig. \ref{fig:dense-cost-reg-net}(f), 
it consists of multiple feature-dense block units, 
each containing a 3D convolution, group normalization, and ReLU activation. 
Outputs from these units are concatenated, 
progressively expanding the receptive field 
and facilitating accurate estimation across multiple depth planes. 
This compact yet effective design significantly 
improves cost volume regularization performance.

Together, these modules form a powerful
pipeline comprising a flexible feature extraction network with global feature aggregation, 
and a robust cost regularization network featuring a densely connected encoder for rich representation learning 
and a feature-dense decoder with an expanding receptive 
field to effectively handle multiple depth planes estimates.
Empirical validations underscore the efficacy of these components, 
collectively driving significant improvements in the MVS pipeline.

\begingroup
\setlength{\tabcolsep}{2pt} 
\begin{table*}[t]
\caption{Quantitative results on Tanks and Temples \cite{Knapitsch2017tnt}. \textbf{Bold} and \underline{underline} represents first and second place, respectively.}
\vspace{-10pt}
\footnotesize
  \begin{center}
    \begin{tabular}{lccccccccc|ccccccc}
    \toprule
    & \multicolumn{9}{c}{\textbf{Intermediate set}} & \multicolumn{7}{c}{\textbf{Advanced set}} \\
    \cmidrule{2-17}
    Method & \textbf{Mean $\uparrow$} & Fam. & Fra. & Hor. & Lig. & M60 & Pan. & Pla. & Tra. & \textbf{Mean $\uparrow$} & Aud. & Bal. & Cour. & Mus. & Pal. & Tem.\\
    \midrule
    COLMAP \cite{Johannes2016pixelwise} & 42.14 & 50.41 & 22.25 & 26.63 & 56.53 & 44.83 & 46.97 & 48.53 & 42.04 & 27.24 & 16.02 & 25.23 & 34.70 & 41.51 & 18.05 & 27.94\\
    R-MVSNet \cite{yao2019recurrent} & 50.55 & 73.01 & 54.56 & 43.42 & 43.88 & 46.80 & 46.69 & 50.87 & 45.25  & 29.55 & 19.49 & 31.45 & 29.99 & 42.31 & 22.94 & 31.10\\
    CasMVSNet \cite{gu2019casmvsnet} & 56.84 & 76.37 & 58.45 & 46.26 & 55.81 & 56.11 & 54.06 & 58.18 &  49.51 & 31.12 & 19.81 & 38.46 & 29.10 & 43.87 & 27.36 & 28.11\\
    AA-RMVSNet \cite{wei2021aa} & 61.51 & 77.77 & 59.53 & 51.53 & 64.02 & 64.05 & 59.47 & 60.85 & 54.90  & 33.53 & 20.96 & 40.15 & 32.05 & 46.01 & 29.28 & 32.71\\
    UniMVSNet \cite{peng2022rethinkingMVS} & 64.36 & 81.20 & 66.43 & 53.11 & 63.46 & \textbf{66.09} & 64.84 & \underline{62.23} & 57.53  & 38.96 & 28.33 & 44.36 & 39.74 & 52.89 & 33.80 & 34.63 \\
    TransMVSNet \cite{ding2022transmvsnet} & 63.52 & 80.92 & 65.83 & 56.94 & 62.54 & 63.06 & 60.00 & 60.20 & 58.67  & 37.00 & 24.84 & 44.59 & 34.77 & 46.49 & 34.69 & 36.62\\
    GBi-Net \cite{mi2022gbi}  & 61.42 & 79.77 & 67.69 & 51.81 & 61.25 & 60.37 & 55.87 & 60.67 & 53.89 & 37.32 & 29.77 & 42.12 & 36.30 & 47.69 & 31.11 & 36.39  \\
    MVSTER \cite{wang2022mvster}  & 60.92 & 80.21 & 63.51 & 52.30 & 61.38 & 61.47 & 58.16 & 58.98 & 51.38 & 37.53 & 26.68 & 42.14 & 35.65 & 49.37 & 32.16 & 39.19 \\
    GeoMVSNet \cite{zhe2023geomvsnet} & 65.89  & 81.64  & 67.53  & 55.78 & 68.02  & 65.49 & 67.19 & \textbf{63.27} & 58.22  & 41.52 & 30.23 & 46.53  & 39.98 & \underline{53.05} & 35.98 & 43.34\\
    MVSFormer \cite{caomvsformer} & 66.37  & 82.06  & \underline{69.34}  & 60.49 & \underline{68.61}  & \underline{65.67} & 64.08 & 61.23 & 59.53  & 40.87 & 28.22 & 46.75  & 39.30 & 52.88 & 35.16 & 42.95 \\
    MVSFormer++ \cite{cao2024mvsformer} & \textbf{67.03}  & \textbf{82.87}  & 68.90  & \textbf{64.21}  & \textbf{68.65}  & 65.00  & \textbf{66.43}  & 60.07  & 60.12  & 41.70  & \underline{30.39}  & 45.85  & 39.35  & \textbf{53.62}  & 35.34  & \underline{45.66} \\
    GoMVS \cite{wu2024gomvs} & \underline{66.44} & 82.68 & 69.23 & \underline{63.19} & 63.56 & 65.13 & 62.10 & 58.81 & \underline{60.80} & \textbf{43.07} & \textbf{35.52} & \underline{47.15} & \textbf{42.52} & 52.08 & \underline{36.34} & 44.82 \\
    GC-MVSNet \cite{vats2023gcmvsnet} & 62.74  & 80.87  & 67.13  &  53.82 & 61.05  & 62.60  & 59.64  &  58.68 & 58.48 & 38.74 & 25.37  &  46.50 & 36.65 & 49.97 & 35.81 & 38.11\\
    \midrule
    GC-MVSNet++ & 66.28  & \underline{82.72}  & \textbf{71.05}  & \underline{58.18} &  65.40 & 65.63 & \underline{64.95} & 61.47 & \textbf{60.87}  & \underline{42.14} & 27.74 & \textbf{47.84}  & \underline{40.24} & 52.32 & \textbf{37.48} & \textbf{47.23}\\ 
    \bottomrule
    \end{tabular}
    \label{table:quantitative-comparison-tanks-and-temples}
    \vspace{-15pt}
\end{center}
\end{table*}
\endgroup

\section{Experiments}\label{sec:experiments}

We evaluate on three datasets of different
complexities. \textit{DTU} \cite{jensen2014dtu} is an indoor dataset
that contains $128$ scenes with $49$ or $64$ views under $7$ lighting
conditions and predefined camera trajectories. We use the same training, validation, and test
splits as MVSNet
\cite{yao2018mvsnet}.  \textit{BlendedMVS} \cite{yao2019blended} is a
large-scale synthetic dataset with $113$ indoor and outdoor scenes. It
has 106 training scenes and $7$ validation scenes. \textit{Tanks and
  Temples} \cite{Knapitsch2017tnt} is collected from a more
complicated and realistic scene, and contains $8$ intermediate and $6$
advanced scenes. DTU and Tanks \& Temples evaluate using point clouds while BlendedMVS
evaluates on depth maps.

\vspace{-3pt}
\subsection{Implementation Details}\label{sec:implementation-details}

Following  general practice \cite{ding2022transmvsnet}, we first train
and evaluate our model on DTU. Then, we finetune on BlendedMVS to
evaluate on Tanks and Temples. For training on DTU, we set the number
of input images to $N=5$ and image resolution to $512 \times 640$. The
depth hypotheses are sampled from $425mm$ to $935mm$ for
coarse-to-fine regularization with the number of plane sweeping depth
hypotheses for the three stages set to 48, 32, and 8. The corresponding depth
interval ratio (DIR) is set as 2.0, 0.8, and 0.4. The model is trained
with Adam \cite{Kingma2014AdamAM} for 9 epochs with an initial
learning rate ($LR_{DTU}$) of 0.001, which decays by a factor of 0.5
once after $8^{th}$ epoch. For the GC-module, we use $M$=8 and set
the stage-wise thresholds $D_{pixel}$ as 1, 0.5, 0.25 and ${D_{depth}}$ as
0.01, 0.005, 0.0025. We use $\alpha$=$\beta$=1 and $\gamma=2$ for all experiments. 
We train our model with a batch size of 2 on 8
NVIDIA RTX A6000 GPUs for about 9 hours.

\subsection{Experimental Performance}\label{sec:experimental-performance}

\noindent \textbf{Evaluation on DTU:}
We generate depth maps with $N$=5 at an input
resolution of $864\times1152$. We slightly adjust the depth interval ratio (DIR) to $1.6,
0.7, 0.3$ to accommodate the resolution change (more on DIR in
Appendix C) and use the Fusibile algorithm \cite{Galliani2015fusibile} for
depth fusion.
Table \ref{Table:DTU-Blended-model-comparison} shows quantitative
evaluations, where accuracy is the mean
absolute distance in $mm$ from the reconstructed point cloud to the
ground truth point cloud, completeness measures the opposite
(more details in Appendix E), and overall is the average of
these metrics indicate the overall performance of the
models. 
We find that GC-MVSNet++ achieves the second best overall and
completeness scores when compared to previous state-of-the-art techniques.
Snapshots of the DTU test set point clouds are shown in the Supplementary Materials.
We find that our model
generates denser and more complete point clouds.

\noindent \textbf{Evaluation on BlendedMVS:}
Unlike DTU and Tanks and Temples, evaluation on Blended MVS is usually measured
as the quality of depth maps, not the quality of point clouds.
First, we finetune our model for 12 epochs with $N$=5, $M$=8, number of depth planes $D$=128, at a 
resolution of $576 \times 768$, with one-tenth the learning rate of DTU ($\frac{1}{10} LR_{DTU}$).

Following evalutation process of Darmon et al. \cite{darmon2021wildMVS}, we generate
Table \ref{Table:DTU-Blended-model-comparison} for a quantitative comparison with other 
methods using three metrics:
Endpoint error (EPE) is the average $L_1$
distance between the estimated and the ground truth depth values, and
$e_1$ and $e_2$ are the ratio of number of pixels with $L_1$ error
larger than 1\textit{mm} and 3\textit{mm}, respectively.
The significant jump in depth estimates corroborates our intuition that providing explicit geometric cues during training
helps the model be mindful about preserving the geometric consistency of a view during inference. 
The addition of dense-CostRegNet and the use of the depth filtration module led to 
significant quantitative and qualitative improvements for  GC-MVSNet++.
error maps and point clouds are shown in Appendices H and I, respectively.).

\begin{figure*}[t]
\begin{center}
    \includegraphics[trim={0 0 4cm 0},clip, width=0.95\textwidth,height=17\baselineskip]{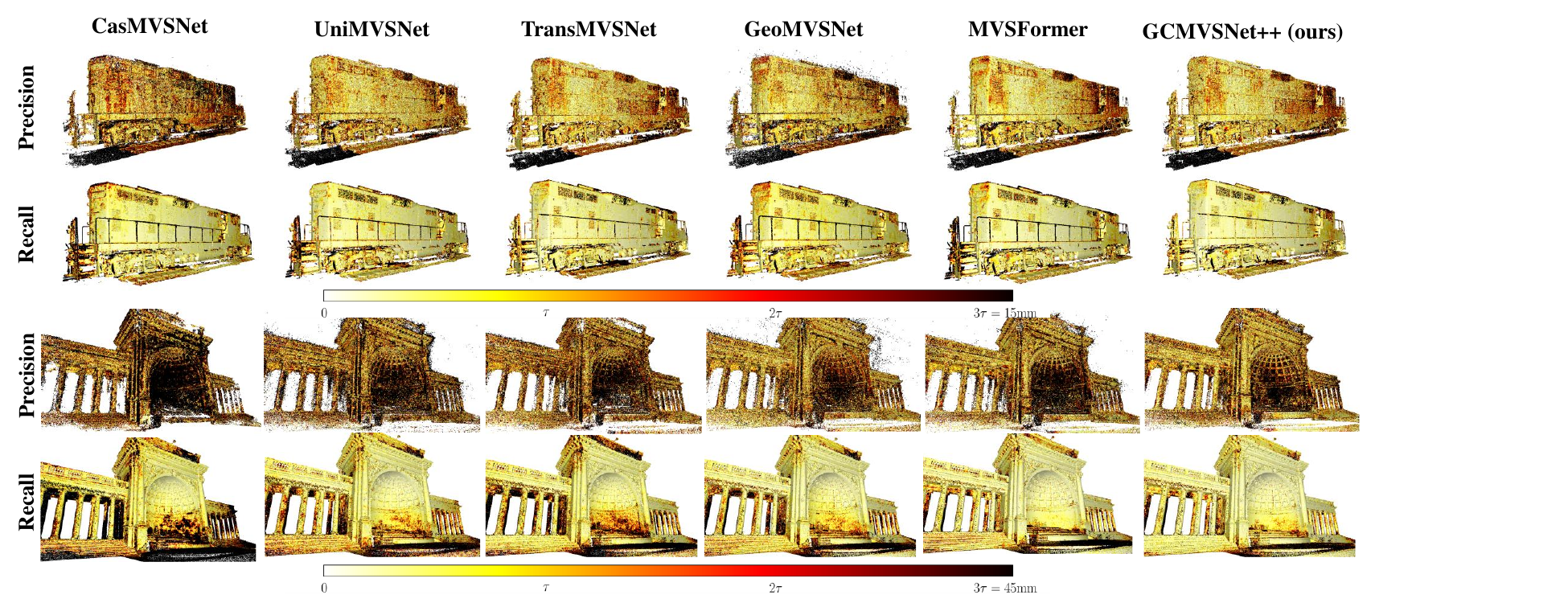}
    \caption{Precision and recall comparison with other recent methods for 
    Train (intermediate set) and Temple (advanced set) scene on Tanks and Temples benchmark. The error plots are collected from 
    the benchmark website.
    $\tau$ is the scene-relevant distance threshold. Darker regions indicate larger errors
    encountered with regard to $\tau$. GC-MVSNet++ shows visual improvements with brighter 
    regions for both metrics.}
    \label{fig:tnt-point-cloud-main-paper}
\end{center}
\end{figure*}

\vspace{2pt}
\noindent \textbf{Evaluation on Tanks and Temples:}
To test the generalization ability of our model, we use Tanks and Temples dataset -- a high-resolution outdoor benchmark.
To adapt to the indoor-to-outdoor change, we first finetune our model on BlendedMVS 
with $N$=7, $M$=10, $D$=192 at $576 \times 768$ resolution with one-tenth the learning rate of DTU ($\frac{1}{10}LR_{DTU}$) 
for $12$ epochs, and evaluate on a greater than 2$\times$ higher resolution of $1080 \times 1920$\footnote{a few scenes have $1080\times2048$ resolution.}.
The camera parameters and neighboring view
selection are used as in R-MVSNet \cite{yao2019recurrent} 
and follow
evaluation steps described in MVSFormer \cite{caomvsformer}.
Table
\ref{table:quantitative-comparison-tanks-and-temples} presents a
quantitative comparison with other methods. GC-MVSNet++ achieves  
second best on the advanced set and has a competitive performance on the intermediate set. 
Fig. \ref{fig:tnt-point-cloud-main-paper} shows point clouds visualizing precision
and recall comparisons for Train (intermediate set) and Temple (advanced set)
with other MVS methods. These plots are downloaded from the Tanks and Temple
benchmark leaderboard. Point clouds are shown in Appendix I.

\subsection{Ablation Study}\label{sec:ablation-study}

We conduct ablation  studies 
to evaluate the importance of the various components of GCMVSNet++.
We provide detailed comparisons for inference time and memory requirements  with 
MVSFormer++, TransMVSNet, CasMVSNet, and GC-MVSNet in Appendix F of the 
supplementary material. 

\vspace{2pt}
\noindent \textbf{$\mathbf{\xi_p}$ Range:}
$\xi_p$ is generated using the mask sum ($mask\_sum$ in Alg. \ref{alg:gc-algorithm}), and  
is the sum of penalties accumulated across the $M$ source views during 
multi-view geometric consistency check. At this stage, its elements take a discrete value 
between $0$ and $M$. Using mask sum as-is leads to a very high penalty
per-pixel, and thus a very high loss value, 
destabilizing the learning process. We control the magnitude 
of the penalty by controlling the range of the per-pixel penalty. 

We explore two different ranges,
$[1,2]$ and $[1,3]$. To generate $\xi_p \in [1,2]$, we divide the mask
sum by $M$ and then add $1$. To
generate $\xi_p \in [1,3]$, we divide the mask by $\frac{M}{2}$
and then add $1$. Table
\ref{table:range-of-per-pixel-penalty} shows the impact of these two
ranges  for $M$=8. Since $\xi_p \in [1,2]$ produces the best results,
we use it for all other experiments.

\vspace{2pt}
\noindent \textbf{Hyperparameters of GC module}
The GC module has two types of hyperparameters, global and local. In this section,
we investigate the effect of these hyperparameters on our results.

The global hyperparameter $M$ is the number of source views across which
the geometric consistency is checked, and is the same for all three
stages (coarse, intermediate, and refinement stages).  For training on
DTU, we vary the value of $M$ while keeping $N$=5, i.e. while the MVS
method uses only $4$ source views to estimate $D_0$, the GC module
checks the geometrical consistency of $D_0$ across $M$ source
views. It is important to note that the first $N$-1 out of $M$ source
views are exactly the same used by GC-MVSNet++ to estimate $D_0$. We
always keep $M \geq N-1$.

Table \ref{table:src views gc check ablation} presents a quantitative
comparison for different values of $M$ and the number of training
iterations required for optimal performance of the model. At $M$=$N-1$= $4$,
i.e. checking geometric consistency across the same number of source
views as used by GC-MVSNet++ to estimate $D_0$, the model performance
significantly improves with a sharp decrease in training iteration requirements, 
as compared to our baseline GC-MVS-base. As we increase the
value of $M$ from $3$ to $10$, the training iterations required by our model
further decrease. We find that at $M=8$, which is twice the number of
source views used by GC-MVSNet++, it achieves its best performance.

The two local hyperparameters, $D_{pixel}$ and $D_{depth}$, are the
stage-wise thresholds applied to generate $PDE_{mask}$ and
$RDD_{mask}$ in Alg.\ref{alg:gc-algorithm}. These are set to smaller values in the later (finer) stages,
providing a stricter
penalty to geometrically inconsistent pixels at finer
resolutions. Table \ref{table:gc-module-hyperparameter} shows the
overall performance of GC-MVSNet++ with a range of different $D_{pixel}$
and $D_{depth}$ thresholds. GC-MVSNet++ performance remains fairly consistent and  achieves its best performance
with $D_{pixel}= 1, 0.5, 0.25$ and $D_{depth}= 0.01, 0.005, 0.0025$. We
use these threshold values throughout the paper.

\begin{table}[t]
\caption{Impact of range of $\xi_p$ during training on DTU with $M$=8, $N$=5. Numbers are generated on DTU evaluation set.}
\setlength{\tabcolsep}{3pt}
  \begin{center}
    {\footnotesize{
\begin{tabular}{cccc}
\toprule
$\xi_p$ Range  & Acc$\downarrow$ & Comp$\downarrow$ & Overall$\downarrow$ \\
\midrule
$[1,3]$ & 0.331 & 0.270 & 0.3005 \\ 
$[1,2]$ & \textbf{0.330} & \textbf{0.260} & \textbf{0.295} \\
\bottomrule
\end{tabular}}}
\label{table:range-of-per-pixel-penalty}
\end{center}

\caption{Quantitative results on DTU evaluation set \cite{jensen2014dtu}. M is the number of source views used by the GC module for checking the geometric consistency of the reference view depth map. The training iteration requirement of the model decreases as M increases.}
\setlength{\tabcolsep}{2pt}
  \begin{center}
    {\footnotesize{
\begin{tabular}{ccccc}
\toprule
Src. Views (M)  & Acc$\downarrow$ & Comp$\downarrow$ & Overall$\downarrow$ & Opt. Epoch\\
\midrule
3 & 0.334 & 0.274  & 0.304 & 14 \\
4 & 0.343 & 0.264  &  0.3035  & 12\\
5 & 0.342 & 0.271  & 0.3065   &  13\\
6 & \textbf{0.326} & 0.271  &  0.298  & 9 \\
7 & 0.332 &  0.270 &  0.301  & 10 \\
\textbf{8} & 0.330 &  \textbf{0.260} &  \textbf{0.295}  & 9 \\
9 & 0.328 &  0.280 & 0.304  & 9 \\
10 & 0.329 & 0.268  &  0.298 & 10 \\

\bottomrule
\end{tabular}}}
\label{table:src views gc check ablation}
\end{center}

\caption{Overall score on the evaluation set of DTU \cite{jensen2014dtu} for different 
values of $D_{depth}$ and $D_{pixel}$. $M$ is fixed at $8$. C, I, and R means Coarse, Intermediate, and Refine stages.}
\setlength{\tabcolsep}{2pt}
  \begin{center}
    {\footnotesize{
\begin{tabular}{ccccccc}
    \toprule
    \multicolumn{3}{c}{$D_{depth}$} & \multicolumn{3}{c}{$D_{pixel}$} & \multirow{2}{*}{Overall$\downarrow$} \\
    \cmidrule{1-6}
    Coarse & Inter. & Refine & Coarse & Inter. & Refine &  \\
    \midrule
    0.04 & 0.03 & 0.02 & 4 & 3 & 2 & 0.302\\
    0.03 & 0.0225 & 0.015 & 3 & 2.25 & 1.5 & 0.302 \\
    0.02 & 0.015 & 0.01 & 2 & 1.5 & 1.0 & 0.298 \\
    0.01 & 0.008 & 0.006 & 1 & 0.8 & 0.6 & 0.303 \\
    \textbf{0.01} & \textbf{0.005} & \textbf{0.0025} & \textbf{1} & \textbf{0.5} & \textbf{0.25} & \textbf{0.295} \\
    0.008 & 0.003 & 0.002 & 0.8 & 0.3 & 0.2 & 0.303 \\
    0.005 & 0.002 & 0.001 & 0.5 & 0.2 & 0.1 & 0.3015\\
    \bottomrule
\end{tabular}
}}
\label{table:gc-module-hyperparameter}
\end{center}
\end{table}

\vspace{2pt}
\noindent \textbf{GC-module as a plug-in:}
Geometric Consistency module is generic and can be integrated into
many different MVS pipelines
To demonstrate this, we test it with
two very different MVS pipelines, TransMVSNet and CasMVSNet.  
CasMVSNet treats depth
estimation as a regression problem, while TransMVSNet treats it as a
classification problem and uses winner-take-all to estimate the final
depth map. We purposefully choose different methods to show that the
GC module can perform well for both types of formulation. We 
compare the architectures of GC-MVSNet++ with TransMVSNet and CasMVSNet
in discussion section. We also compare it with MVSFormer and MVSFormer++. 

Table \ref{table:performance-comparison-with-GC-module} presents the results,
showing the impact
of adding the GC module, the deformable feature extraction network (shown as \textit{other} modifications in the table), and depth filtration
in the original method pipeline. To observe the absolute impact of adding these
modifications, we do not modify the original
pipelines in any other way. We do not include our proposed dense-CostRegNet or Deformable feature extraction network
in the original methods as it would 
enhance their learning capabilities. We focus on the extent of improvement 
in the original method with the integration of the GC module and depth filtration preprocessing step. 
We observe in the table that applying only the \textit{other}
modification leads to degradation in performance for both methods. It
indicates that the \textit{other} modification helps in
stabilizing the training process and promoting reproducibility but
has no significant impact on the performance of the model on its
own. We also observe a sharp increase in model
performance and a decrease in training iteration requirements after integrating
our GC module into the original pipeline. With GC, training the
CasMVSNet pipeline requires only $11$ epochs 
instead of $16$ epochs, while TransMVSNet (with GC module) requires only $8$ epochs
instead of $16$ epochs. This corroborates our
hypothesis that multi-view geometric consistency significantly reduces training computation
because it accelerates the geometric cues learning. Applying the depth filtration 
preprocessing along with the GC module further improves the performance of both methods.
Eliminating the erroneous ground truth pixels from the learning process through 
filtration provides more consistent geometric cues learning. 

\begin{table}[t]
    \caption{The impact of GC as a plug-in module on CasMVSNet and TransMVSNet on DTU \cite{jensen2014dtu}.
Apart from other changes, including a GC module alongside depth filtration (DF)
in the original methods boosts the performance. 
$L_1$ and D-FEN indicate $L_1$ loss, Focal loss \cite{Lin2017FocalLoss}, and 
Deformable Feature Extraction Network, respectively. 
DF is the depth filtering pre-processing step to remove erroneous ground truth values.}
\setlength{\tabcolsep}{2pt}
  \begin{center}
    {\footnotesize{
\begin{tabular}{llccccc}
\toprule
  Methods & Loss & DF & Other & GC & Overall$\downarrow$ & Epoch\\
\cmidrule{2-7}
 \multirow{4}{*}{CasMVSNet}& $L_1$ & $\times$ & $\times$ & $\times$ & 0.355 & 16\\
  & $L_1$ & $\times$ & $\checkmark$ & $\times$ & 0.357 & 16\\
  & $L_1$ & $\times$ & $\times$ & $\checkmark$ & 0.335 & 11 \\
  & $L_1$ & $\checkmark$ & $\times$ & $\checkmark$ & \textbf{0.330} & \textbf{11} \\
\cmidrule{2-7}
\multirow{4}{*}{TransMVSNet} & FL & $\times$  & $\times$ & $\times$ & 0.305 & 16 \\
 & FL& $\times$ & $\checkmark$ & $\times$ & 0.322 & 16 \\
 & FL& $\times$ & $\times$ & $\checkmark$ & 0.303 & 8 \\
& FL& $\checkmark$ & $\times$ & $\checkmark$ & \textbf{0.297} & \textbf{8} \\
\bottomrule
\end{tabular}
}}
\label{table:performance-comparison-with-GC-module}
\end{center}

    \caption{Stages of performance improvement of GC-MVSNet++ with different modifications on DTU \cite{jensen2014dtu}. 
It uses Cross-entropy loss \cite{wei2021aa} for training. DF, D-FEN, GC, GFA, and DenseCostReg indicate depth
filtering, Deformable-Feature Extraction Network, Geometric Consistency module, Global Feature Aggregation
and Dense Cost Regularization network, respectively.}
\setlength{\tabcolsep}{2pt}
  \begin{center}
    {\footnotesize{
\begin{tabular}{lcccccc}
\toprule
Model & D-FEN & GFA & GC & DF & CostRegNet & Overall$\downarrow$\\
\midrule
GC-MVS-Base & $\times$ & $\times$ & $\times$ & $\times$ & $\times$ & 0.332 \\
GC-MVS-Base & $\times$ & $\checkmark$ & $\times$ & $\times$ & $\times$ & 0.334 \\
GC-MVS-Base  & $\checkmark$ & $\checkmark$ & $\times$& $\times$ & $\times$ & 0.328 \\
GC-MVS-Base  & $\times$ & $\times$ & $\checkmark$& $\times$ & $\times$ & 0.298 \\
GC-MVS-Base  & $\checkmark$ & $\checkmark$ & $\checkmark$& $\times$ & $\times$ & 0.295\\
GC-MVS-Base  & $\checkmark$ & $\checkmark$ & $\checkmark$& $\checkmark$ & $\times$ & 0.291\\
GC-MVSNet++  & $\checkmark$ & $\checkmark$ & $\checkmark$ & $\checkmark$ & $\checkmark$ & \textbf{0.2825}\\
\bottomrule
\end{tabular}
}}
\label{table:stages-of-gcmvsnet++}
\end{center}
\end{table}

\vspace{2pt}
\noindent \textbf{Stages of GC-MVSNet++:}
Table \ref{table:stages-of-gcmvsnet++} shows
different stages of development of GC-MVSNet++.  
GC-MVS-base uses the TransMVSNet pipeline (without the feature matching transformer and 
adaptive receptive field modules) with cross-entropy loss
and performs much worse than original TransMVSNet which uses focal loss.
Including the global feature aggregation module with the basic feature extraction network 
performs roughly similar. 
But including \textit{D-FEN} (Deformable-Feature Extraction Network)
modifications slightly improves the overall performance indicating 
better feature quality with deformable convolution for global feature aggregation. However, D-FEN does not 
reduce the training iteration requirements on its own, see Table. \ref{table:performance-comparison-with-GC-module}. Only after applying the GC module, 
independently and with D-FEN, we see a significant reduction 
in training iteration requirements and a significant improvement in the overall 
accuracy metric. This clearly shows the significance of multi-view multi-scale geometric consistency 
checks in the GC-MVSNet++ pipeline. Including the DF preprocessing step further aids the GC module by 
eliminating erroneous ground truth values from the training loop and provides more robust 
geometry cues across multiple source views. Finally, adding dense-CostRegNet with 
all other modifications improves the cost regularization process with its
advanced architecture and dense identity connection, leading to a significant improvement.

\section{Discussion}
\label{sec:discussion}

We provide a comparison of our model with recent state-of-the-art models like 
MVSFormer, MVSFormer++, TransMVSNet and CasMVSNet.

\vspace{3pt}
\noindent \textbf{GC-MVSNet++ vs. MVSFormer and MVSFormer++:}
MVSFormer focuses on using 
state-of-the-art pre-trained models for feature extraction and extensively use 
transformers in all its components. MVSFormer++ further adds multiple transformer modules 
in the MVS pipeline to leverage global feature aggregation in cost volume regularization. 
GC-MVSNet++ uses only one global feature aggregation 
module after feature extraction for global context enhancement before cost volume creation.
It focuses on enforcing geometry-based learning across multiple views 
and on effective utilization of dense connections
to enhance the cost regularization network. 

\vspace{3pt}
\noindent \textbf{GC-MVSNet++ vs. TransMVSNet:}
TransMVSNet uses regular 2D
convolution-based FPN (with batch-norm) for feature extraction and
employs adaptive receptive field (ARF) modules with deformable layers
after feature extraction. It trains using focal loss
\cite{Lin2017FocalLoss}. GC-MVSNet++ replaces the combination of regular
FPN and ARF modules with weight-standardized deformable FPN
(with group-norm) for feature extraction. It also uses a novel densely 
connected CostRegNet for cost volume regularization. It trains with a combination 
of cross-entropy  
loss and geometric consistency penalty for accelerated learning.

\vspace{3pt}
\noindent \textbf{GC-MVSNet++ vs. CasMVSNet:}
CasMVSNet \cite{gu2019casmvsnet} proposes a coarse-to-fine regularization
technique. It uses a regular 2D convolution-based FPN for feature
extraction, generates variance-based cost volume, and employs depth
regression to estimate $D_0$. The only similarity with our model is
that we also use coarse-to-fine regularization.

\section{Conclusion}

In this paper, we introduce GC-MVSNet++, an enhanced learning-based 
MVS pipeline that explicitly models the geometric consistency of 
reference depth maps across multiple source views during training. 
Our approach incorporates a novel dense-CostRegNet with two 
distinct modules, simple dense-module which is effective for encoding and 
feature-dense-module, which is designed to leverage dense connection for effective decoding and precise estimation 
of reference view depth maps. Ours is the first few method to integrate multi-view, multi-scale geometric consistency 
checks into the training process. We demonstrate that 
the GC module is versatile and can be integrated with 
other MVS methods to enhance their learning. 
Through extensive experiments and ablation studies, 
we highlight the advantages of GC-MVSNet++. 
This work shows how to  blend 
traditional geometric techniques and modern machine 
learning methods, to achieve more accurate 
and reliable 3D reconstructions from multiple images.

\textbf{Acknowledgement:} This work was supported by Electronics and Telecommunications Research Institute (ETRI) grant funded by the government of the Republic of Korea [25ZC1110, The research of the basic media $\cdot$ contents technologies].

\appendix

\begin{figure*}[ht]
    \centering
   \includegraphics[width=1\textwidth,height=15\baselineskip]{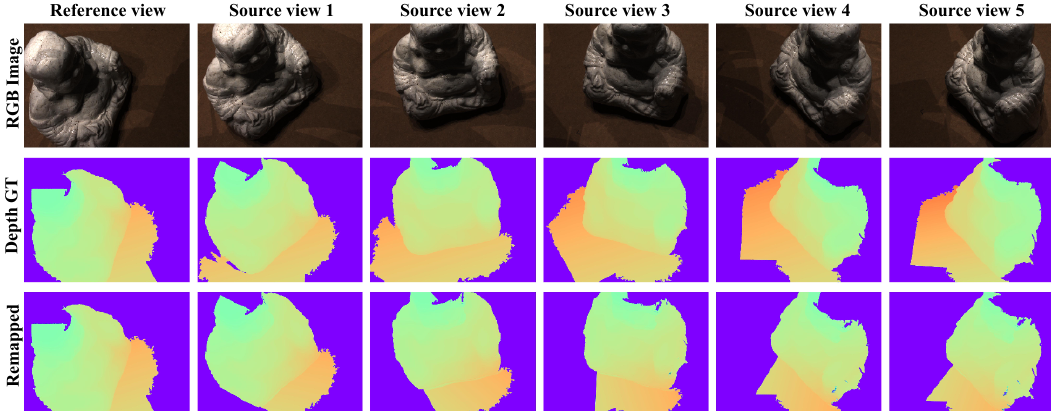}
   \caption{The first row illustrates the selection of the $M$ closest source 
  images for a given reference image. The middle row displays the corresponding ground truth depth maps, while the last 
  row shows the remapped source depth maps, achieved by projecting the reference view’s x-y coordinates onto the source view. 
  During remapping, any additional pixels from the source views are discarded. The remapped depths are then back-projected 
  onto the source view to generate a mask. This reference view mask is applied to the per-pixel penalty to limit the penalties. 
  The resulting final $\xi_p$ is presented in Fig. 3 of the paper. All depth maps are displayed within their respective view masks.}
    \label{fig:occlusion-handling}
\end{figure*}

\section{Occlusion and its impact}

Modeling pixel occlusion in multi-view settings presents a significant challenge, 
particularly when reasoning about pixels whose corresponding 3D points are occluded in other views. 
This issue becomes more pronounced when penalties are assigned to all such pixels, 
as in the proposed multi-view geometric consistency (GC) module. The GC module evaluates 
the geometric consistency of each pixel across multiple source views and imposes penalties 
for inconsistencies. However, penalizing occluded pixels and combining these penalties with 
depth errors negatively affects the training process. In our initial experiments, 
we observed that this approach caused the loss to explode during training, 
meaning the loss values increased as the model continued to train.

Our investigation reveals that incorrect penalties applied to occluded pixels dominated the loss during training. 
To address this, we implemented a series of steps that made our method more robust to this issue. 
First, we selected the closest source view images, as defined by MVSNet \cite{yao2018mvsnet}, 
to reduce the likelihood of occluded pixels. The first row of Fig. \ref{fig:occlusion-handling} 
illustrates this source view selection for a given reference view. Using the nearest 
views minimizes the number of potentially occluded pixels.

Second, during forward-backward reprojection, we remap the source view depth map 
based on the x-y coordinate projections from the reference view to the source view, 
and then back-project these remapped values to the reference view (as described in Alg. 2 of the paper). 
The last row of Fig. \ref{fig:occlusion-handling} shows the remapped source view depth maps. 
During this remapping process, occluded and extraneous pixels from the source view are discarded, 
ensuring that only valid pixels are back-projected. This step effectively 
manages extreme cases of occlusion and additional visible pixels.

Finally, after generating the per-pixel penalty, we apply a binary mask from the reference view 
to exclude any pixels not part of the scene (see Fig. 3 in the paper). 
This combination of steps significantly reduces the impact of incorrect 
penalties and stabilizes the training process.

\section{Geometric Consistency Module}

\begin{figure}[t]
\begin{center}
   \includegraphics[width=0.5\linewidth]{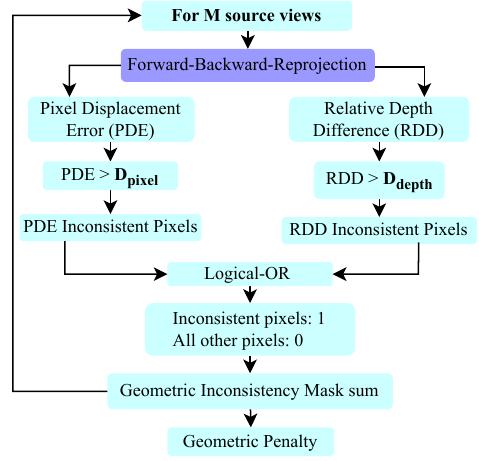}
\end{center}
   \caption{GC module flow-chart for consistency check.}
\label{fig:gc-module-flow-chart}
\end{figure}

We describe the steps of the geometric consistency (GC) module in Fig. \ref{fig:gc-module-flow-chart}. 
At each stage, the geometric consistency of the estimated depth map is checked across $M$ source views. 
For each source view, we perform the forward-backward-reprojection of 
estimated depth map to reason about geometric inconsistency of pixels (described in Alg. 2). 
In this three-step process, first, we warp each pixel $P_{0}$ of a reference view depth map $D_{0}$ to its $i^{th}$ 
neighboring source view to obtain corresponding pixel $P^{'}_{i}$. 
Then, we back-project $P^{'}_{i}$ into 3D space and finally, 
we reproject it to the reference view as $P^{"}_{0}$ 
using $c_0$. $D_0$, $D^{'}_{P^{'}_{i}}$ and $D^{"}_{P^{"}_{0}}$ represents 
depth value of pixels associated with $P_0$, $P^{'}_{i}$ and $P^{"}_{0}$ \cite{Hartley2012Multi-view-geometry}. 
With $P^{"}_{0}$ and $D^{"}_{P^{"}_{0}}$, 
we calculate pixel displacement error (PDE) and relative depth difference (RDD). 
After taking logical-OR between PDE and RDD, we assign value $1$ to all inconsistent pixels and zero to all other pixels. 
The geometric inconsistency mask sum is generated over $M$ source views and averaged to generate per-pixel penalty $\xi_p$.

\section{Depth Interval Ratio (DIR)}

\begin{table}[t]
\caption{The performance of GC-MVSNet on evaluation set of DTU \cite{jensen2014dtu} with change in stage-wise DIR (depth interval ratio).}
  \begin{center}
    {\small{
\begin{tabular}{clccc}
\toprule
$\xi_p$ Range  & Stage-wise DIR  & Acc$\downarrow$ & Comp$\downarrow$ & Overall$\downarrow$ \\
\midrule
$[1,3]$ & 2.0, 0.8, 0.40 & 0.338 & 0.269 & 0.3035 \\
$[1,3]$ & 2.0, 0.7, 0.35 & 0.343 & \textbf{0.264} & 0.3035 \\
$[1,3]$ & 2.0, 0.7, 0.30 & 0.331 & 0.27 & 0.3005 \\ 
$[1,3]$ & 1.6, 0.7, 0.30 & \textbf{0.329} & 0.271 & \textbf{0.300} \\
\bottomrule
\end{tabular}}}

\label{table:dir}
\end{center}
\end{table}

DIR directly impacts the separation of two hypothesis planes at pixel level. For a given stage, the pixel-level depth interval is calculated as product of $DIR_{stage}$ and \textit{depth interval} (DI). The value of DI is calculated using \textit{interval scale} and a constant value provided in DTU camera parameter files. 

Following the trend of modern learning-based methods \cite{xu2019multiscale, gu2019casmvsnet, yao2018mvsnet, YU2021AAcostvolume, ding2022transmvsnet, peng2022rethinkingMVS, Cheng2019USCNet}, we train our model on $512 \times 640$ resolution and test on $864 \times 1152$ resolution.  To adjust for the pixel-level depth interval caused by the increase in resolution, we explore different DIR values for testing on DTU. We train our model with stage-wise DIR $2.0, 0.8, 0.4$ ($DIR_{train}$).  such that the refine stage pixel-level depth interval is same as the provided \textit{interval scale} value of $1.06$. Table \ref{table:dir} shows DIR values for evaluation on DTU, we only explore smaller values than $DIR_{train}$ to compensate for the increase in resolution. GC-MVSNet achieves its optimal performance at DIR $1.6, 0.7, 0.3$ with $\xi_p \in [1,3]$, $DIR_{test}$. We use the same $DIR_{train}$ and $DIR_{test}$ with $\xi_p \in [1,2]$.

\section{Depth Map Fusion Methods}

The quality of point clouds depends heavily on depth fusion methods and their hyperparameters. Following the recent learning-based methods \cite{ding2022transmvsnet, peng2022rethinkingMVS, gu2019casmvsnet}, we also use different fusion methods for DTU and Tanks and Temples dataset. For DTU, we use Fusibile \cite{Galliani2015fusibile} and for Tanks and Temples, we use Dynamic method \cite{ding2022transmvsnet, wei2021aa}.

The fusibile fusion method uses three hyperparameters, disparity threshold, probability confidence threshold, and consistency threshold. The disparity threshold defines the upper limit of disparity for points to be eligible for fusion. The probability confidence threshold defines the lower limit of confidence above which points are eligible for fusion. The consistency threshold mandates that the eligible points be geometrically consistent across as many source views. During the fusion process, only those points that satisfy all three conditions are fused into a point cloud. 

The dynamic fusion method uses only two hyperparameters, probability confidence threshold and consistency threshold. Both these hyperparameters have exact same function as in the Fusibile method. The disparity threshold is not provided by the user, it is dynamically adjusted during the fusion process. 

\section{Accuracy and Completeness Metrics}

\begin{figure}[t]
\begin{center}
   \includegraphics[width=0.5\linewidth]{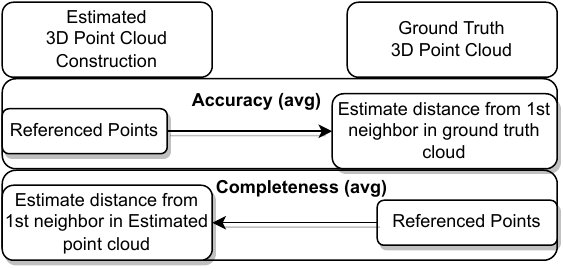}
\end{center}
   \caption{The process of calculating accuracy and completeness for DTU \cite{jensen2014dtu} point cloud evaluation.}
\label{fig:acc-comp-diagram}
\end{figure}

Accuracy and completeness are two metrics used with the DTU \cite{jensen2014dtu} dataset. Fig. \ref{fig:acc-comp-diagram} shows the process of calculation. Accuracy is the average of the distance of the first neighbor from the predicted point cloud to the ground truth point cloud. It only considers the points which are below the maximum threshold for the distance. For completeness, the same process is repeated but with ground truth as a referenced point cloud, i.e. the average of the distance of the first neighbor from the ground truth point cloud to the predicted point cloud. 

\section{Inference Memory Comparison}

\begin{table}[t]
\caption{Illustration of model memory during the inference phase of various models at 
$864 \times 1152$ and $1152 \times 1536$ resolutions. We use information from MVSFormer++ paper 
to fill out CasMVSNet and TransMVSNet information.}
\vspace{-5pt}
\setlength{\tabcolsep}{2pt}
  \begin{center}
    {\footnotesize{
\begin{tabular}{llcc}
\toprule
Methods & Resolution & Depth Interval & Memory (MB) \\
\midrule
\multirow{4}{*}{MVSFormer++} & \multirow{2}{*}{864 $\times$1152} & 32-16-8-4 & 4873\\
&  & 64-32-8-4 & 5025\\
\cmidrule{2-4}
& \multirow{2}{*}{1152 $\times$1536} & 32-16-8-4 & 5964\\
&  & 64-32-8-4 & 6753\\
\cmidrule{2-4}
\multirow{2}{*}{CasMVSNet} & 864 $\times$1152 & \multirow{2}{*}{48-32-8} &  4769\\
& 1152 $\times$ 1536 &  & 6672\\
\cmidrule{2-4}
\multirow{2}{*}{TransMVSNet} & 864 $\times$ 1152 & \multirow{2}{*}{48-32-8} &  3429\\
& 1152 $\times$1536 &  & 6320\\
\cmidrule{2-4}
\multirow{2}{*}{GC-MVSNet} & 864 $\times$1152 & \multirow{2}{*}{48-32-8} & 2787 \\
& 1152 $\times$ 1536 &  & 4831\\
\cmidrule{2-4}
\multirow{2}{*}{GC-MVSNet++} & 864 $\times$1152 & \multirow{2}{*}{48-32-8} & 5221 \\
& 1152 $\times$ 1536 &  & 8193\\
\bottomrule
\end{tabular}}}
\label{table:memory-comparison}
\end{center}
\end{table}

The memory consumption analysis in Table \ref{table:memory-comparison} demonstrates 
that while our proposed \emph{GC-MVSNet++} incurs increased memory usage compared to lighter architectures 
like \emph{GC-MVSNet}, it remains competitive with transformer-based state-of-the-art 
methods such as \emph{MVSFormer++}. This additional memory overhead results 
from the incorporation of dense connectivity in the cost volume regularization module, 
which significantly enhances representational flexibility and multi-scale information propagation. 
Crucially, this design choice leads to competitive depth estimation accuracy 
and robust geometric consistency, clearly justifying the increased memory footprint. 
Our model achieves a favorable balance, offering accuracy comparable to transformer-based 
methods without the added complexity and overhead of transformer layers.

\section{Use of Existing Assets}

We use PyTorch to implement GC-MVSNet. It is based on CasMVSNet \cite{gu2019casmvsnet} and TransMVSNet \cite{ding2022transmvsnet}. These two methods heavily borrow code from the PyTorch implementation of MVSNet \cite{yao2018mvsnet}. 

We use preprocessed images and camera parameters of DTU \cite{jensen2014dtu} dataset from the official repository of MVSNet \cite{yao2018mvsnet} and R-MVSNet \cite{yao2019recurrent}. We follow \cite{darmon2021wildMVS} for training and testing on BlendedMVS \cite{yao2019blended}. For Tanks and Temples \cite{Knapitsch2017tnt} evaluation, we use images and camera parameters as used in R-MVSNet \cite{yao2019recurrent}.

\section{$e_1$ Error Comparison on BlendedMVS}

\begin{figure*}[t]
\begin{center}
    \includegraphics[width=1\textwidth]{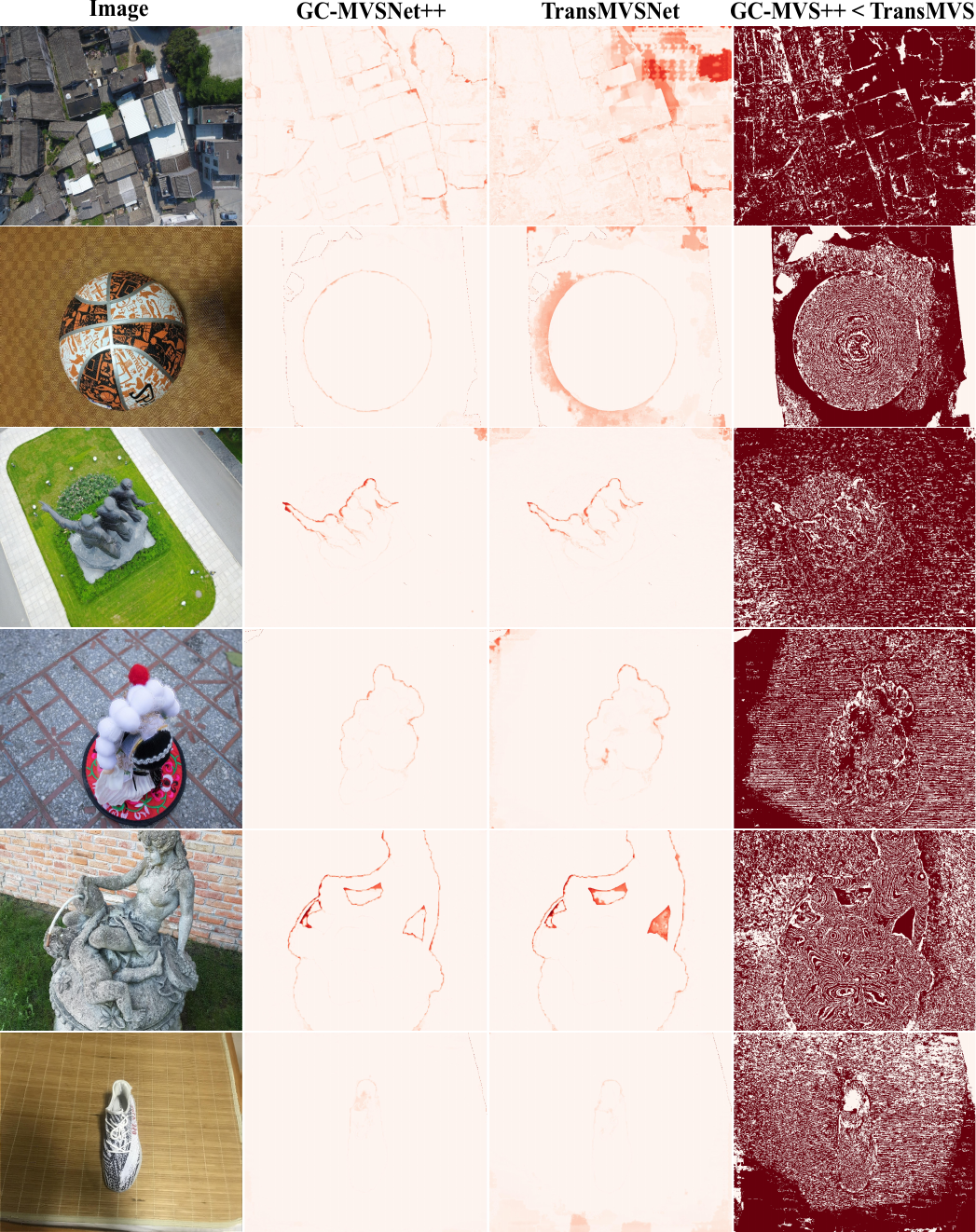}
    \vspace{-10pt}
    \caption{EPE error comparison of GC-MVSNet++ with TransMVSNet on validation set of BlendedMVS dataset. First column 
    shows the image of the scene, second and third columns show their respective EPE error maps (darker is higher error), and last column shows 
    the binary map (0,1) of all the points where GC-MVSNet++ has smaller EPE-error as compared to TransMVSNet.}
    \label{fig:e1-error-plots}
\end{center}
\end{figure*}

In this section, we discuss the $e_1$-error plots for the GC‑MVSNet++ and TransMVSNet methods. 
A quantitative comparison table is provided in the main paper. 
We argue that training MVS methods with geometry‑based guidance not only accelerates 
the optimization process but also significantly improves depth‑estimation consistency. 
Figure \ref{fig:e1-error-plots} shows the error-map comparison with TransMVSNet. 
The first column displays images from a BlendedMVS test‑set scenes; 
the second and third columns show the absolute-error plots (darker red is higher error) for GC‑MVSNet++ 
and TransMVSNet, respectively. Although there are visual differences in these error maps, 
the fourth column reveals more detail: it presents a binary map (0 or 1) of all points 
where GC‑MVSNet++’s error is lower than TransMVSNet’s. This clearly demonstrates 
the advantage of geometry‑guided training over non‑geometry‑guided methods in MVS. 
While most test‑scene images exhibit visible improvements in their error-maps, 
these gains do not translate linearly into the final point‑cloud quality.

\section{Point Clouds}

\begin{figure*}[ht]
\begin{center}
    \includegraphics[width=0.85\textwidth]{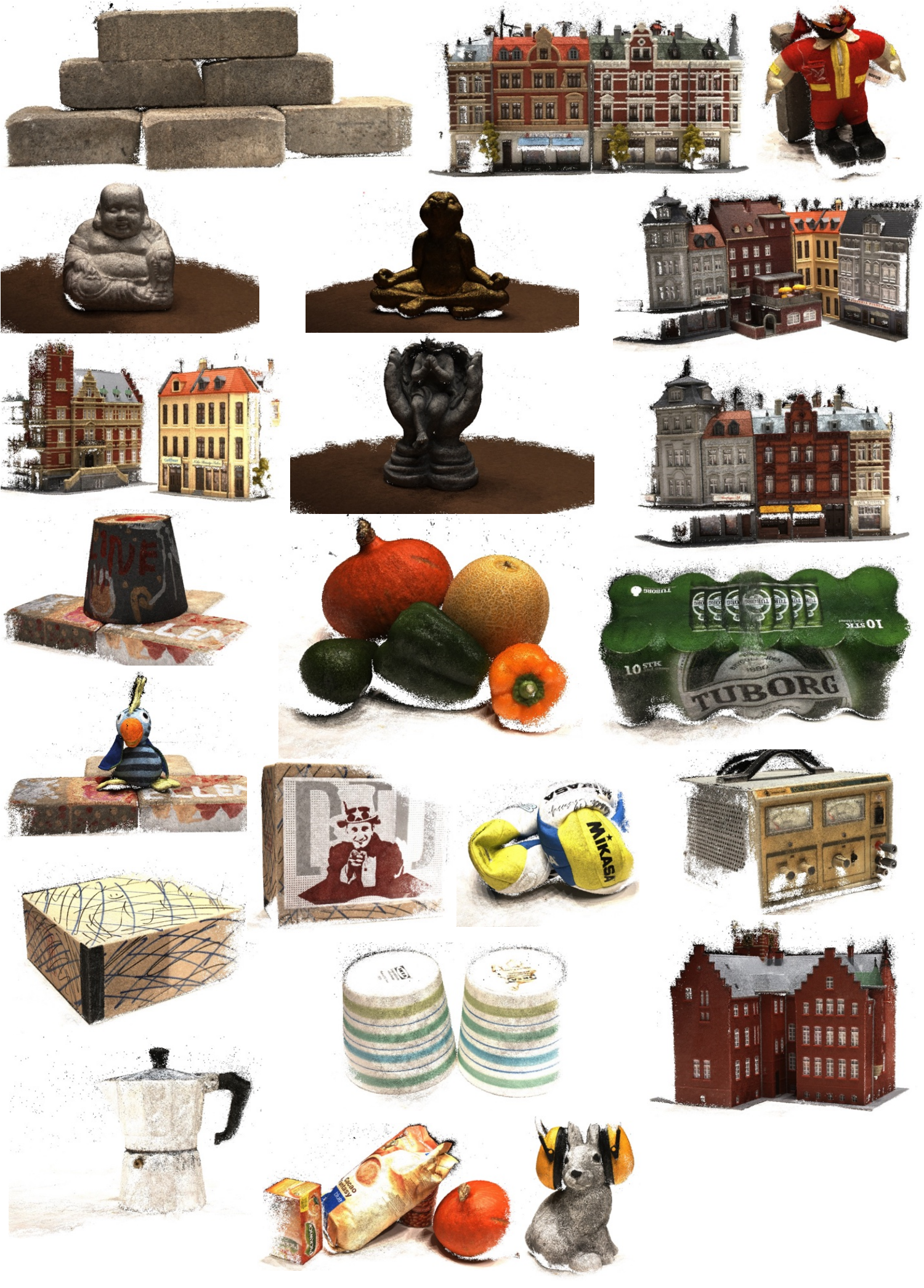}
    \caption{Point clouds reconstructed using GC-MVSNet for all scenes from DTU \cite{jensen2014dtu} evaluation set.}
    \label{fig:dtu-point-clouds}
\end{center}
\end{figure*}

\begin{figure*}[t]
\begin{center}
    \includegraphics[width=0.98\textwidth]{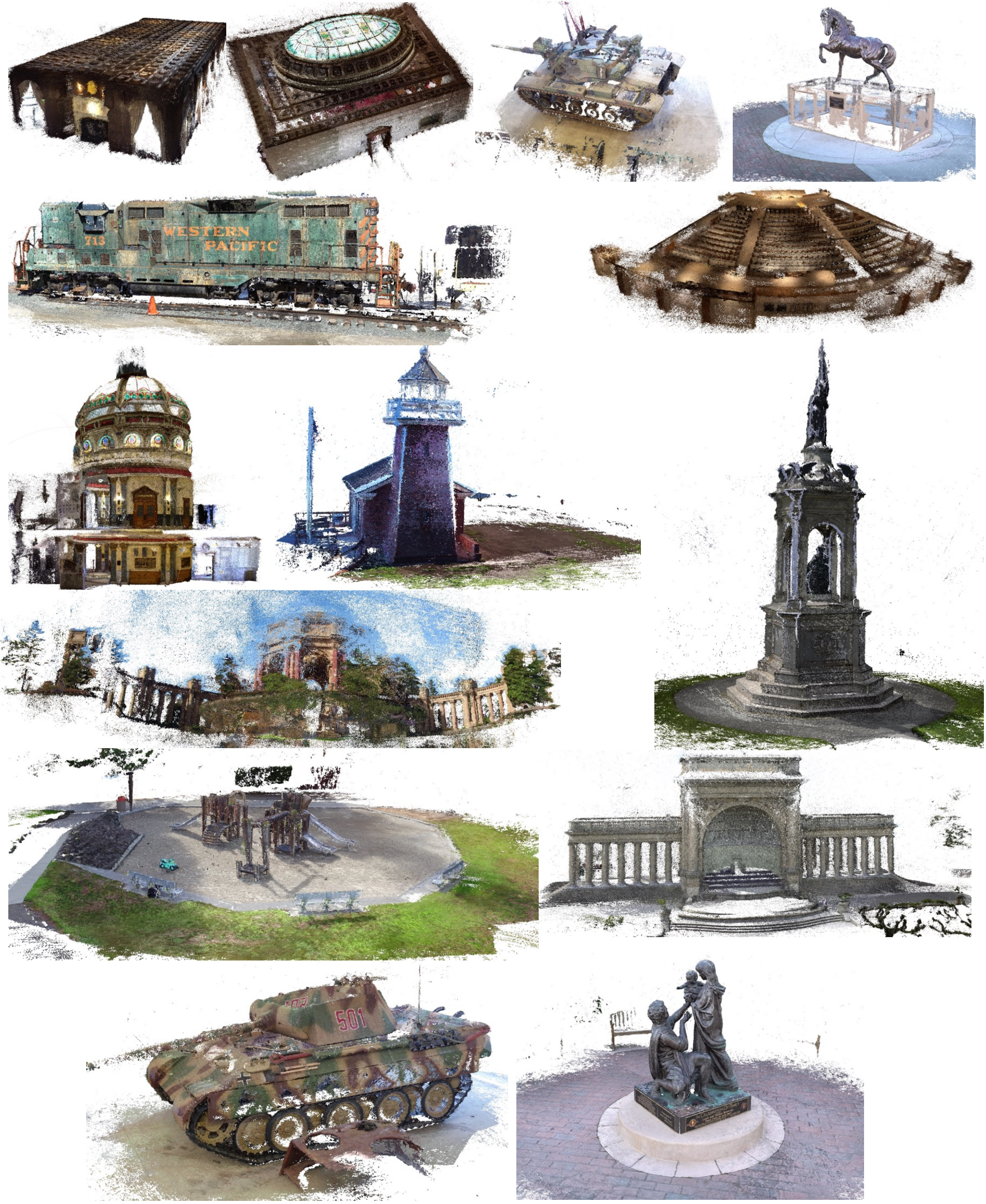}
    \caption{Point clouds reconstructed using GC-MVSNet for all scenes from Tanks and Temples \cite{Knapitsch2017tnt} intermediate and advanced set.}
    \label{fig:tnt-point-clouds}
\end{center}
\end{figure*}

\begin{figure*}[t]
\begin{center}
    \includegraphics[width=0.95\textwidth]{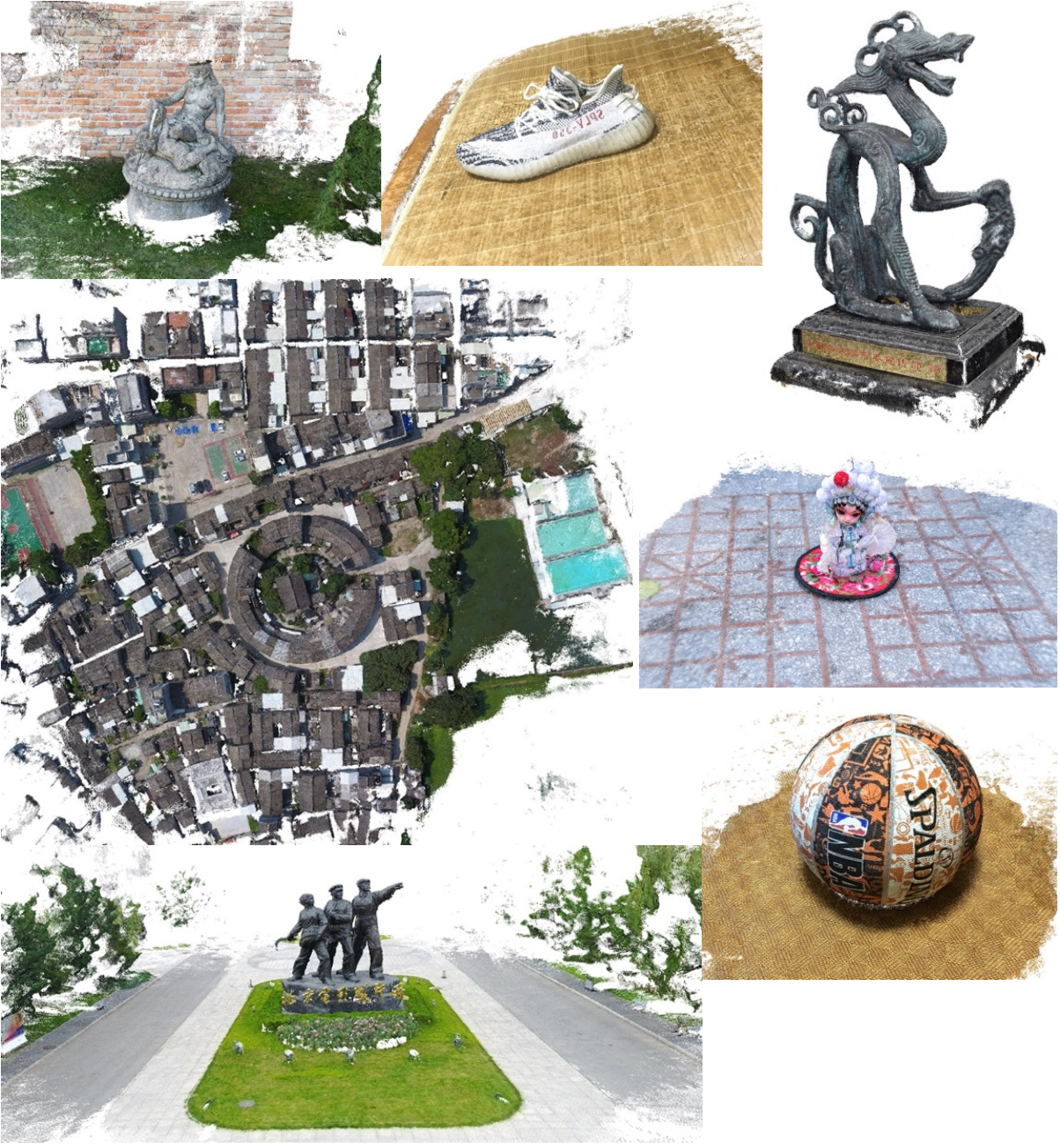}
    \caption{Point clouds reconstructed using GC-MVSNet for all scenes from BlendedMVS \cite{yao2019blended} evaluation set.}
    \label{fig:blended-point-clouds}
\end{center}
\end{figure*}

In this section, we show all evaluation set points clouds reconstructed using GC-MVSNet on DTU \cite{jensen2014dtu}, Tanks and Temples \cite{Knapitsch2017tnt} and BlendedMVS \cite{yao2019blended} datasets. Fig. \ref{fig:dtu-point-clouds}, \ref{fig:tnt-point-clouds} and \ref{fig:blended-point-clouds} show all evaluation set point clouds from DTU, Tanks and Temples and BlendedMVS, respectively.

\bibliographystyle{unsrt}
\bibliography{paper,reza}

\end{document}